\documentclass[11pt,a4paper]{article}
\usepackage{times,latexsym}
\usepackage{url}
\usepackage[T1]{fontenc}

\usepackage[acceptedWithA]{tacl2021v1}
\usepackage[T1]{fontenc}
\usepackage{graphicx}
\usepackage{amssymb,amsmath,amsthm,enumitem}
\usepackage{bbm}
\usepackage{cleveref}
\usepackage{bbm}
\usepackage{xcolor}
\definecolor{skyblue}{RGB}{0,102,204}
\definecolor{orangebrown}{RGB}{204,102,0}
\definecolor{darkgray}{rgb}{0.66, 0.66, 0.66}
\usepackage{xspace,mfirstuc,tabulary}

\usepackage{graphicx}
\usepackage{booktabs}
\usepackage{url}

\newif\iftaclinstructions
\taclinstructionsfalse % AUTHORS: do NOT set this to true
\iftaclinstructions
\renewcommand{\confidential}{}
\renewcommand{\anonsubtext}{(No author info supplied here, for consistency with
TACL-submission anonymization requirements)}
\newcommand{\instr}
\fi

\iftaclpubformat % this "if" is set by the choice of options

\else

\fi

\title{\textcolor{skyblue}{Ev\textsuperscript{2}}\textcolor{orangebrown}{R}: \textcolor{skyblue}{Ev}aluating  \textcolor{skyblue}{Ev}idence \textcolor{orangebrown}{R}etrieval in Automated Fact-Checking}

\author{Mubashara Akhtar\textsuperscript{1,2}, Michael Schlichtkrull\textsuperscript{3} \and {\bf Andreas Vlachos\textsuperscript{4}} \\[5pt]
        \textsuperscript{1}Department of Computer Science, ETH Zurich \hspace{0.5em} 
        \textsuperscript{2}ETH AI Center\\ 
        \textsuperscript{3}School of Electronic Engineering and Computer Science, Queen Mary University of London\\
        \textsuperscript{4}Department of Computer Science and Technology, University of Cambridge\\[5pt]
        \texttt{mubashara.akhtar@ai.ethz.ch}\\ 
        \texttt{m.schlichtkrull@qmul.ac.uk}\\ 
        \texttt{av308@cam.ac.uk}}

\date{}

\begin{document}

% set space before and after equations
\setlength{\abovedisplayskip}{6pt}
\setlength{\belowdisplayskip}{6pt}

\maketitle
\begin{abstract}

Current automated fact-checking (AFC) approaches typically evaluate evidence either implicitly via the predicted verdicts or through exact matches with predefined closed knowledge sources, such as Wikipedia. However, these methods are limited due to their reliance on evaluation metrics originally designed for other purposes and constraints from closed knowledge sources. 
% Recent advances in natural language generation (NLG) evaluation offer new possibilities for evidence assessment.
In this work, we introduce \textbf{\textcolor{skyblue}{Ev\textsuperscript{2}}\textcolor{orangebrown}{R}}
%, a weighted scorer that 
which
combines the strengths of reference-based evaluation and verdict-level proxy scoring. Ev\textsuperscript{2}R jointly assesses how well the evidence aligns with the gold references and how reliably it supports the verdict, addressing the shortcomings of prior methods.
We evaluate Ev\textsuperscript{2}R against three types of evidence evaluation approaches: reference-based, proxy-reference, and reference-less baselines. Assessments against human ratings and adversarial tests demonstrate that Ev\textsuperscript{2}R consistently outperforms existing scoring approaches in accuracy and robustness. It achieves stronger correlation with human judgments and greater robustness to adversarial perturbations, establishing it as a reliable metric for evidence evaluation in AFC.\footnote{Code is available at \href{https://github.com/mubasharaak/fc-evidence-evaluation}{https://github.com/mubasharaak/fc-evidence-evaluation}.}

% prompt-based scorers, particularly those leveraging LLMs and reference evidence, outperform traditional evaluation approaches.

\end{abstract}

% \iftaclpubformat
\section{Introduction}
\label{sec:introduction}

To decide the truthfulness of a claim, professional fact-checkers search for, retrieve and
analyse evidence~\citep{graves2018a}. Their goal is not only to assess a claim's accuracy, but also to present the evidence and fact-checking steps transparently, helping readers better understand and trust the fact-checking process~\citep{graves2017}.%\blfootnote{\textsuperscript{*} This work was partially done during Mubashara's research visit at the University of Cambridge.}

% https://docs.google.com/drawings/d/1JTKmROMBB9I9gomxn7-vvr-kfROShkuEHXVK5uSN_Ec/edit?usp=sharing
\begin{figure*}[h!]
    \centering
    \resizebox{1.0\textwidth}{!}{%
    \includegraphics{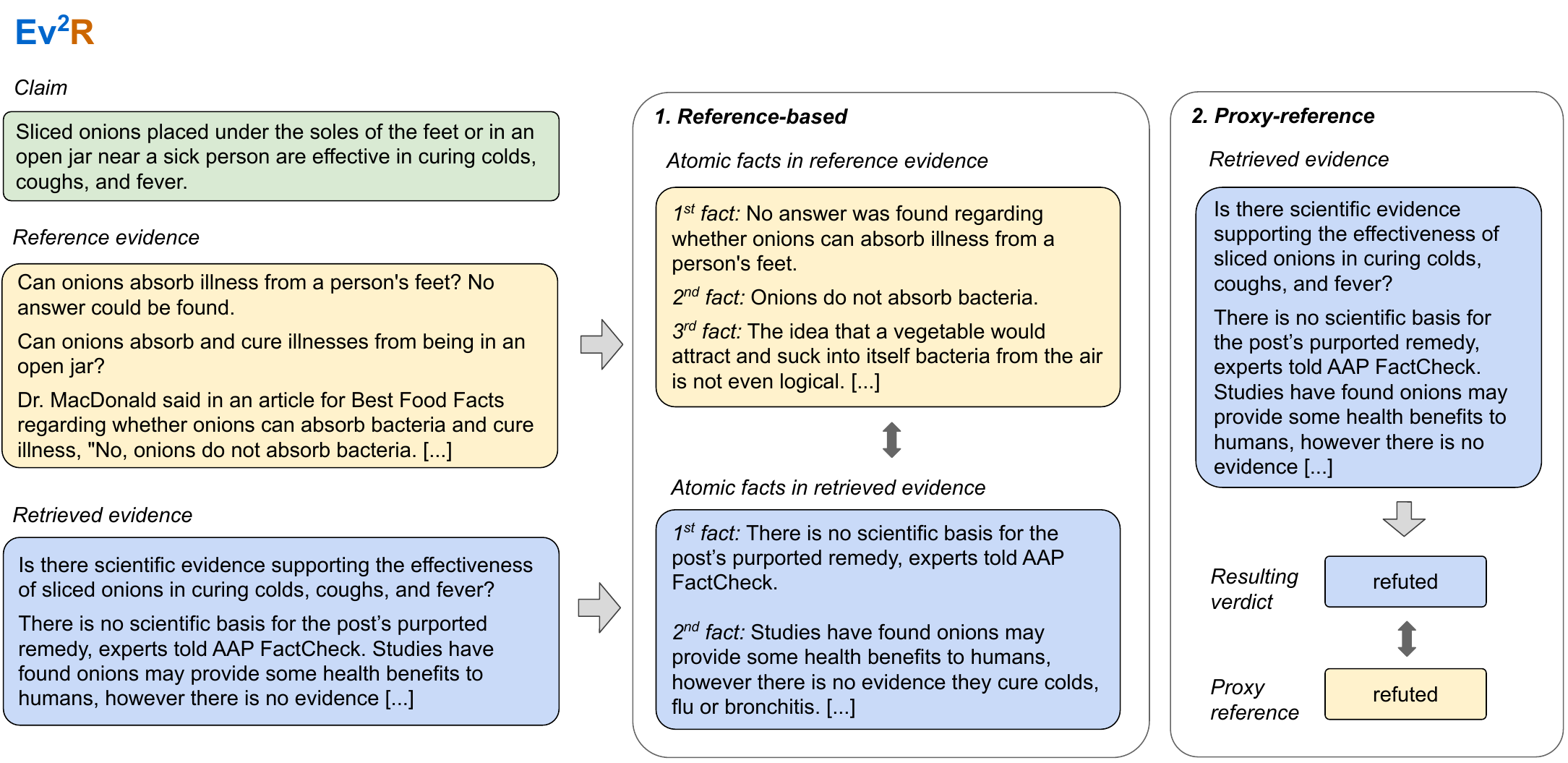}
    }
    \vspace{-2em}
    \caption{Overview of the proposed \textbf{\textcolor{skyblue}{Ev\textsuperscript{2}}\textcolor{orangebrown}{R}} scorer, including the  reference-based (middle) and proxy-reference (right) components. For reference-based evaluation, evidence is first decomposed into atomic facts before evaluation. Proxy-reference scoring uses the verdict label as proxy to assess retrieved evidence.}
    \label{fig:overview_framework}
    % \vspace{-1em}
\end{figure*}

% Challenges related to evidence evaluation: done in the past like this.
Past research has primarily employed three approaches to evaluate evidence in Automated Fact-Checking (AFC). First, some methods assess evidence implicitly through the predicted verdict, assuming that accurate evidence leads to correct verdicts~\citep{ShahiN20, HorneKA18, Barron-CedenoES18}. This overlooks the possibility that retrieved evidence may lead to a different verdict from the gold evidence without directly assessing the evidence itself.
Second, some approaches restrict evidence retrieval to predefined, closed knowledge sources like Wikipedia, comparing the retrieved evidence to reference evidence from the same source and considering only exact matches correct. While common in AFC benchmarks~\citep{thorne-etal-2018-fever, jiang-etal-2020-hover, sathe-etal-2020-automated, norregaard-derczynski-2021-danfever}, this method may dismiss valid evidence not annotated as part of the reference, ignoring the broader variety of trusted open-web sources, such as government statistics.
Lastly, in recent datasets requiring evidence retrieval from the entire internet~\citep{schlichtkrullGV2023, chen-etal-2024-complex}, evaluation often uses token-matching metrics like METEOR~\citep{banerjee-lavie-2005-meteor}. However, these metrics are highly sensitive to surface form differences and fail to account for alternative evidence paths. For example, both \textit{``Where did South Africa rank in alcohol consumption? In 2016, South Africa ranked ninth out of $53$ African countries.''} and \textit{``What's the average alcohol consumption per person in South Africa? $7.1$ liters.''} can both be valid steps on the way to establish relative levels of alcohol consumption between South Africa and other countries.

In this paper, we introduce Ev\textsuperscript{2}R, a weighted scorer for evidence evaluation in AFC that combines two complementary scoring signals: $(i)$ a reference-based scoring component, which assesses retrieved evidence through comparison with reference evidence data, and $(ii)$ a second component, which evaluates evidence using the annotated verdict label as proxy reference.
The reference-based scoring enhances interpretability by clearly demonstrating evidence accuracy (precision) and completeness (recall) relative to annotations.
% However, since the goal of fact-checking systems is to support correct verdict prediction, 
Moreover, for a given claim, multiple sets of evidence or alternative reasoning paths may lead to the same (correct) verdict. As such, looking at verdicts can give (noisy) signal for evidence evaluation.
Hence, we incorporate verdict-level alignment via a proxy-reference scoring component into Ev\textsuperscript{2}R. 
Taking both into account, the Ev\textsuperscript{2}R score
% provides a balanced and comprehensive assessment of evidence for AFC.
reflects both the factual correctness and completeness of the retrieved evidence with respect to the reference, as well as its alignment with the verdict.

We evaluate Ev\textsuperscript{2}R across three AFC datasets (i.e., AVeriTeC~\citep{schlichtkrull-etal-2024-automated}, FEVER~\citep{thorne-etal-2018-fever}, and VitaminC~\citep{schuster-etal-2021-get}) and compare it against evidence scorers used in previous research, including reference-based, proxy-reference, and reference-less baselines. 
AVeriTeC contains real-world claims with web-based evidence, reflecting practical challenges in fact-checking. FEVER and VitaminC are widely-used large-scale benchmarks, each providing multiple evidence sets per claim, sometimes with conflicting verdicts (e.g., one supporting and another refuting the same claim). Together, these datasets allow us to evaluate Ev\textsuperscript{2}R across diverse settings.

We assess Ev\textsuperscript{2}R based on its correlation with human ratings and its robustness to adversarial tests with perturbed evidence. Our results demonstrate that the Ev\textsuperscript{2}R scorer outperforms baselines in terms of correlation with human ratings across all datasets. Additionally, adversarial stress tests reveal that Ev\textsuperscript{2}R is more robust to common perturbations, such as redundancy and noise, compared to traditional metrics used in previous research. These findings highlight that combining reference- and verdict-level scoring provides a more reliable approach for evidence evaluation in AFC.

\section{Related Work}
\label{sec:related_work}

% \todo{check links for related work:}
% https://aclanthology.org/2023.gem-1.25.pdf (suggested by Michael)
% ...

\subsection{Evidence-based AFC}

While earlier AFC datasets evaluate evidence implicitly through the predicted veracity label~\citep{guo-etal-2022-survey, akhtar-etal-2023-multimodal}, more recent approaches require explicit evaluation to ensure that system predictions are grounded in evidence.

There are several challenges in the automated retrieval and evaluation of evidence for evidence-based AFC. First, evidence-based AFC datasets released in recent years contain purpose-built claims derived from Wikipedia, such as FEVER~\citep{thorne-etal-2018-fever}, Hover~\citep{jiang-etal-2020-hover}, WikiFactCheck~\citep{sathe-etal-2020-automated}, and ChartCheck~\citep{akhtar-etal-2024-chartcheck}. These datasets restrict potential evidence to Wikipedia pages from which the claims were derived. While this restriction is unrealistic, it allows the use of simple, conditional scoring functions for evidence evaluation. For example, the FEVERScore~\citep{thorne-etal-2018-fact} assess the verdict conditioned on retrieving the correct evidence.

To create a more realistic setting, recent datasets use evidence retrieved from the entire internet, such as AVeriTeC~\citep{schlichtkrullGV2023} and MultiFC~\citep{augenstein-etal-2019-multifc}. While their claims and evidence setup better align with real-world fact-checking, this complicates evaluation. Retrieved evidence can no longer be evaluated using exact matches on document identifiers or paragraphs. 
% Multiple documents may contain the correct information, with variations in style or format (e.g., evidence in form of a table or as text).
% 
Second, for a given claim, there may be more than one correct set of evidence pieces, as different reasoning chains with varying evidence can be used to verify the claim. For example, to contradict the claim, ``Dr.\ Fauci stated during a recent press statement that wearing masks in outdoor spaces is not recommended,'' evidence can state that the speaker was not Dr.\ Fauci or prove that the content of the statement itself is incorrect.

Finally, correct evidence may be wrongly interpreted as incorrect if it became available after the dataset's creation and is therefore not included in the set of annotated evidence within the dataset.

\subsection{Evaluation Approaches for Natural Language Generation}
\label{ssec:eval_nlg}

To evaluate evidence in the context of AFC, related work on evaluating natural language generation (NLG) systems can serve as inspiration due to the similarly open-ended nature of the task, e.g., a system can return multiple equally valid summaries for the same input text which should be evaluated against a single gold reference summary~\citep{Celikyilmaz20}. While earlier NLG evaluation methods used string-matching metrics to compare generated and reference text based on heuristics~\cite{Celikyilmaz20, lin-2004-rouge, papineni-etal-2002-bleu}, more recent approaches, involve training language models for evaluation~\citep{ZhangKWWA20, sellam-etal-2020-bleurt} or use large language models' (LLMs) conditional language generation capabilities~\citep{FuNJL23}. 
% 
% In this section, we provide an overview of NLG metrics, categorized into reference-based, proxy-reference, and reference-less metrics, in alignment with the scorers we use as baselines in this work.

\paragraph{Reference-based evaluation}
%As the name already indicates, 
Reference-based metrics evaluate the generated text $\hat{x}$ by comparing it to a gold reference $x$: 
$$
y = f(x, \hat{x}),
$$
where the scoring function $f(.)$ can be conceptualized in various ways, for example, as a non-parametric function based on $n$-grams or a similarity function based on conceptualized representations of text.  
Earlier NLG metrics used $n$-grams to compare generated and reference text based on different heuristics such as string overlap, string distance or lexical diversity~\cite{Celikyilmaz20, lin-2004-rouge, papineni-etal-2002-bleu}. In recent years, various LLM-based metrics were proposed, which also serve as inspiration for Ev\textsuperscript{2}R. 
BLEURT, NUBIA, BERTScore, and MoverScore are all evaluation metrics that use contextualized representations (such as BERT~\citep{devlin-etal-2019-bert} embeddings) to assess the similarity between generated and reference text. BLEURT is a learned metric trained on synthetic sentence pairs~\citep{SaiMK23}, NUBIA~\citep{kane-etal-2020-nubia} follows a three-step approach for similarity scoring, BERTScore uses token-level similarity with greedy matching\citet{ZhangKWWA20}, and MoverScore allows partial matching between n-grams/words for more flexible evaluation~\citep{zhao-etal-2019-moverscore}.

\paragraph{Proxy-reference evaluation}
This family of metrics uses a proxy for evaluation and does not directly compare the generated text ($\hat{x}$) to the annotated reference text ($x$).
In previous research they have been applied if the gold references were not available for evaluation or to focus on specific evaluation criteria (e.g., factuality) for which appropriate proxies are defined as we highlight below~\citep{nema-khapra-2018-towards, durmus-etal-2020-feqa, huang-zhang-2021-evaluation}.
Multiple proxy-based metrics for NLG evaluation were proposed in the past~\citep{eyal-etal-2019-question, scialom-etal-2019-answers}. 
Previous works use proxy references like questions and answers, relational tuples, or entailment labels to evaluate generated text. For example, \citet{durmus-etal-2020-feqa} assess if relevant information from the input text is present in the generated text, while other works use relational tuples to check if the generated text matches factual information from the reference text~\citep{GoodrichRLS19}. Additionally, entailment labels are used to detect hallucinations in generated text, based on the idea that a summary should entail its source document~\citep{JiLFYSXIBMF23}.

\paragraph{Reference-less evaluation}
This family of metrics evaluates generated text without the availability of any (proxy) reference. 
More recently, the capabilities of autoregressive LLMs (e.g., GPT$4$~\citep{GPT4}) have been used for reference-less evaluation.
\citet{FuNJL23} define evaluation as a text generation problem. Their proposed GPTScore framework is based on the idea that LLMs are more likely, i.e. with higher conditional generation probability, to produce the generated text (e.g., a summary) if it is of high quality and fulfills given specifications based on the input document. 
FactScore~\citep{min-etal-2023-factscore} is a recent LLM-based framework for reference-less evaluation of factuality in LLM-generated text. To evaluate the factuality of generated text, it breaks down the text into atomic facts and verifies if they are supported by the knowledge sources. The overall score corresponds to the ratio of supported facts. 

\section{\textcolor{skyblue}{Ev\textsuperscript{2}}\textcolor{orangebrown}{R}: A Weighted Scorer for Evidence Evaluation}
\label{sec:scorers}

This section introduces Ev\textsuperscript{2}R, a weighted scorer for evaluating evidence in the context of AFC. The scorer combines two distinct but complementary parts: a reference-based component, which decomposes the comparison between retrieved and reference evidence into precision and recall (see \Cref{ssec:ref_based}); and a proxy-based component, which assesses alignment between the predicted and reference verdict labels, where the verdict label acts as a proxy for evaluating evidence accuracy (see \Cref{ssec:proxy-ref}). In \Cref{ssec:weighted_scorer}, we provide an overview of the score calculation, integrating both components.

\subsection{Reference-based}
\label{ssec:ref_based}

The reference-based scorer $\mathcal{S}_{\text{ref-based}}$ compares the retrieved and reference evidence by decomposing them into atomic facts (more details below) and evaluating their alignment using a precision and a recall score:
\[
\mathcal{S}_{\text{ref-based}}(\hat{E}, E) = \left( s_{\text{prec}}, s_{\text{recall}} \right)
\]

The precision score $s_{\text{prec}}$ measures how accurately the retrieved atomic facts align with the reference, while $s_{\text{recall}}$ evaluates the completeness of the retrieved facts compared to the reference.
Inspired by \citet{min-etal-2023-factscore}, our scorer first splits the retrieved evidence $\hat{E}$ and reference evidence $E$ into atomic facts, $A_{\hat{E}}$ and $A_E$ respectively (see the middle box in \Cref{fig:overview_framework}). 
In \Cref{fig:overview_framework} we demonstrate the extraction of atomic facts from evidence. We decompose the following evidence 
\textit{`Is there scientific evidence supporting the effectiveness of sliced onions in curing colds, coughs, and fever? There is no scientific basis for the post’s purported remedy, experts told AAP FactCheck. Studies have found onions may provide some health benefits to humans, however there is no evidence [...]'} into two facts, 
first, \textit{`There is no scientific basis for the post’s purported remedy, experts told AAP FactCheck.'}, 
and second, \textit{`Studies have found onions may provide some health benefits to humans, however there is no evidence they cure colds, flu or bronchitis. [...]'}

The reference-based evaluation component assesses the factual consistency of a predicted evidence by systematically comparing it to the reference evidence.\footnote{\Cref{fig:prompt_reference_based_component} in the appendix shows the exact prompt we use for extracting and verifying facts.} First, the scorer breaks down both the predicted and reference evidence into atomic facts, where each fact is a separate sentence. Then, each fact from the predicted evidence is evaluated to verify whether it is directly supported by any fact in the reference evidence, and vice versa, i.e., each reference fact is checked for support in the predicted evidence. We instruct the scorer to rely on the provided texts, disallowing any external sources or background knowledge. The final output summarizes the total number of facts in each evidence (i.e.\ reference and retrieved) and how many of them are supported, offering a measure of factual alignment.

Based on the extracted and scored facts, the scorer computes precision and recall scores to evaluate evidence quality: precision reflects how many retrieved evidence facts are supported by the reference evidence, %(i.e., accuracy),
while recall measures how many reference evidence facts are round in the retrieved evidence. %(i.e., completeness).}
 % by assessing whether each atomic fact $a_{\hat{E}} \in A_{\hat{E}}$ of the retrieved evidence is supported by the reference evidence $E$. 
 We specify the precision score $s_{\text{prec}}$ as the ratio of facts supported by the reference evidence:
$$
s_{\text{prec}} = \frac{1}{|A_{\hat{E}}|} \sum_{a_{\hat{E}} \in A_{\hat{E}}} \mathbbm{I}[a_{\hat{E}} \text{ supported by } E]
$$

The scorer iterates over each fact ($a_{\hat{E}} \in A_{\hat{E}}$), applying an indicator function ($\mathbbm{I}[a_{\hat{E}} \text{ supported by } E]$) that returns $1$ if the fact $a_{\hat{E}}$ is supported by the reference evidence $E$ and $0$ otherwise.
Similarly, the recall score captures whether each atomic fact of the reference evidence ($a_E \in A_E$) is supported by the retrieved evidence, evaluating the extend to which the retrieved evidence covers the reference evidence:
$$
s_{\text{recall}} = \frac{1}{|A_E|} \sum_{a_E \in A_E} \mathbb{I}[a_E \text{ supported by } \hat{E}]
$$

In addition to precision and recall, we also provide an aggregated $F_1$ score ($s_{f_1}$). 
While the reference-based scorer enables fine-grained reference-based evaluation, it may underestimate retrieved evidence when it is correct but different from the reference set. The proxy component, we introduce in \Cref{ssec:proxy-ref}, helps stabilise evaluation in such cases by assessing the stance of the retrieved evidence against the reference verdict.

We evaluate the reference-based component with several LLMs as backbone models: GPT$4o$, Gemini $1.5$ Pro, Gemini $1.5$ Flash, and Llama 3.1 $70$B. We provide further details on the models in \Cref{sec:evaluation} and the instruction prompts in the appendix (see Figures \ref{fig:prompt_reference_less_component}, \ref{fig:prompt_proxy_reference_component}, and \ref{fig:prompt_reference_based_component}).

\subsection{Proxy-reference}
\label{ssec:proxy-ref}

The proxy-reference scoring component of Ev\textsuperscript{2}R uses the verdict label $y$ as a proxy to assess the retrieved evidence $\hat{E}$. As shown in the right box of \Cref{fig:overview_framework}, this component compares the predicted verdict label to the reference label, i.e., \textit{refuted}.
Proxy-reference scoring uses a fine-tuned DeBERTa language model~\citep{HeLGC21} as backbone to predict the verdict label $y$ serving as a proxy for evidence evaluation. 
% The integrated DeBERTa model was previously trained on multiple relevant benchmarks, i.e., MNLI~\citep{williams-etal-2018-broad}, Fever-NLI~\citep{nie2019combining}, Adversarial-NLI~\citep{nie-etal-2020-adversarial}, LingNLI~\citep{parrish-etal-2021-putting-linguist} and WANLI~\citep{liu-etal-2022-wanli}. To fine-tune the model for the task of automated fact-checking, we use the training sets of several AFC benchmarks: FEVER~\citep{thorne-etal-2018-fever}, VitaminC~\citep{schuster-etal-2021-get}, Hover~\citep{jiang-etal-2020-hover}, and AVeriTeC~\citep{schlichtkrullGV2023}.

The proxy-reference scoring component builds on a DeBERTa-v3 model~\citep{HeLGC21} finetuned to predict verdict labels for claim-evidence pairs. We initialised the scorer from a checkpoint pretrained on a diverse set of NLI datasets, i.e., MNLI~\citep{williams-etal-2018-broad}, FEVER-NLI~\citep{nie2019combining}, Adversarial-NLI~\citep{nie-etal-2020-adversarial}, LingNLI~\citep{parrish-etal-2021-putting-linguist}, and WANLI~\citep{liu-etal-2022-wanli}, which provide a foundation for assessing factual inference.
To adapt the model for evidence evaluation in the context of automated fact-checking, we fine-tuned it further on AFC datasets, including the training sets of FEVER, VitaminC, HOVER, and AVeriTeC. The training was conducted using the HuggingFace Trainer API, mixed-precision (fp16) and an initial learning rate of $1\text{e}^{-6}$. The model was trained for three epochs, a warmup phase of fifty steps, and evaluated every $10k$ steps using the micro-F1 score on held-out evidence sets to select the best checkpoint.

Given a claim $c$, a retrieved evidence set $\hat{E}$, and a reference label $y \in \mathcal{Y}$, the proxy scoring component $\mathcal{S}_{\text{proxy}}$ assigns a confidence score $s_{\text{proxy}}$ to the label $y$, reflecting the model's confidence in the predicted verdict:
\begin{align*}
s_{\text{proxy}} &= \mathcal{S}_{\text{proxy}}(c, y, \hat{E}) \\
                 &= \text{softmax}(\mathbf{z})_y = \frac{e^{z_y}}{\sum_{y' \in \mathcal{Y}} e^{z_{y'}}}
\end{align*}

where $\mathbf{z}$ is the vector of logits over all possible veracity labels $\mathcal{Y}$. 
The score $s_{\text{proxy}} \in [0,1]$ is computed as the softmax of the logits $\mathbf{z}$, where the index $y$ corresponds to the reference label, and the softmax function transforms the logits into a probability distribution over the possible labels.
The resulting softmax distribution quantifies the model's confidence.

% The proxy score $s_{\text{proxy}}$ is computed by applying the softmax function to the model's predicted logits $\mathbf{z} = \mathbf{S}(c, \hat{E})$, and selecting the probability assigned to the reference label $y \in \mathcal{Y}$. This score $s_{\text{proxy}}$ is derived by transforming the model-predicted output logits into probabilities using softmax. 
% The resulting probability distribution allows the scorer to quantify its certainty in the different labels potentially available (e.g., support, refute, and not enough information). The scorer's confidence in the reference label $y$ is used as the output score $s_{proxy} \in [0, 1]$. Hence, evidence increasing confidence in the reference label is rated higher accordingly.

\subsection{Final Weighted Scorer}
\label{ssec:weighted_scorer}

To compute the final Ev\textsuperscript{2}R score, we use the outputs of both previously introduced components: the reference-based scorer $\mathcal{S}{\text{ref}}(\hat{E}, E)$, which yields precision and recall scores ($s{\text{prec}}$ and $s_{\text{recall}}$), and the proxy scorer $\mathcal{S}_{\text{proxy}}(c, y, \hat{E})$, which estimates the confidence in the label $y$.
We first compute the $F_1$ score from the precision and recall outputs of $\mathcal{S}_{\text{ref}}$. Then we calculate the final weighted score as a combination of the resulting $F_1$ score and the $s_{\text{proxy}}$. Specifically, we use a weighting factor $\alpha = 0.5$ to equally balance both components:
\[
s_{\text{\textcolor{skyblue}{Ev\textsuperscript{2}}}\text{\textcolor{orangebrown}{R}}} = \alpha \cdot \left( \frac{2 \cdot s_{\text{prec}} \cdot s_{\text{recall}}}{s_{\text{prec}} + s_{\text{recall}}} \right) + (1 - \alpha) \cdot s_{\text{proxy}}
\]

This ensures that the final score reflects both the factual correctness and completeness of the retrieved evidence with respect to the reference, as well as its alignment with the verdict.

\section{Evaluation of \textcolor{skyblue}{Ev\textsuperscript{2}}\textcolor{orangebrown}{R}}
\label{sec:evaluation}

This section provides an overview of the systematic evaluation approaches conducted as part of the meta-evaluation of Ev\textsuperscript{2}R, which involves three different data sources.
First, using adversarial tests generated through perturbing evidence data (see \Cref{ssec:checklist}).
Second, evidence obtained from AFC systems submitted to the AVeriTeC shared task competition~\citep{schlichtkrull2024automatedverificationtextualclaims} in early 2024. (see \Cref{ssec:human_eval}).
Third, evidence pairs extracted from the AFC benchmarks FEVER~\citep{thorne-etal-2018-fever} and VitaminC~\citep{schuster-etal-2021-get}. We discuss the data sources in more detail below. 
Finally, in \Cref{ssec:baselines}, we give an overview of baseline scorers and models used for evaluation. 

\subsection{Evaluation with Adversarial Tests}
\label{ssec:checklist}

We evaluate the scorers by assessing their robustness against adversarial perturbations, including both semantics-altering and semantics-preserving changes. Specifically, we perform adversarial stress tests, including both semantics-altering and semantics-preserving perturbations, such as redundancy, noise insertion, and completeness alterations.
Inspired by \citet{ribeiro-etal-2020-beyond}, we generated large-scale test sets by perturbing either the claim or evidence data. 
We included both semantics-preserving and semantics-modifying tests (see \Cref{table:adversarial_test_dimensions} in the appendix). For semantics-preserving tests, we change the claim and/or evidence such that the resulting evaluation score should not be affected, for example, by introducing small typos.
These tests assess the scorers' robustness to semantically equivalent changes in the evidence data.
On the other hand, we also create semantics-modifying tests by changing the evidence such that its meaning alters and subsequently scores should drop, for example, by removing a significant part of the evidence. 
To create tests with automated perturbations, we use the AVeriTeC test set as the basis.

% Our tests focus on the seven different dimensions (see \Cref{table:adversarial_test_dimensions} in the appendix), the first two categories alter the semantics of the initial evidence, while the remaining ones preserve the evidence's meaning. 

\subsection{Evaluation with Human Ratings}
\label{ssec:human_eval}

\paragraph{Data source}

The AVeriTeC dataset involves retrieving evidence from the web and assessing the veracity of real-world claims previously checked by professional fact-checkers. It addresses limitations in prior AFC benchmarks through focusing on real-world claims and annotated evidence in question-answer format. 
% Originally consisting of 4,568 claims from fact-checking organizations, the dataset was expanded with an additional test set of 1,215 newer claims for the shared task challenge~\citep{schlichtkrull-etal-2024-automated}. 
The AVeriTeC shared task attracted a total of $21$ submissions of fact-checking systems~\citep{schlichtkrull-etal-2024-automated}.
Participating systems were required to provide URLs for their retrieved evidence as well as verdict labels categorized as supported, refuted, not enough evidence, or conflicting evidence/cherrypicking. The organisers provided a public knowledge store with documents to avoid reliance on costly search APIs.

In collaboration with the AVeriTeC organisers, we obtained a subset of evidence across all submitted systems, equally representing both high- and low-scoring systems. After ranking the systems according to their performance on the shared task, we categorise them in four performance groups each containing approximately $25\%$ of submitted systems to allow comparative analysis: Low score (below $0.16$), mid-low score (between $0.16$ and $0.22$), mid-high score (between $0.22$ and $0.435$), and high score ($0.435$ or higher). This classification enabled systematic examination of system performance across different scoring levels. Using stratified sampling, we selected $560$ claims to ensure an even distribution across the dataset.

\paragraph{Human ratings and agreement}
The shared-task participants were then asked to evaluate these samples.\footnote{Find details on the annotation platform in the appendix.} Thirteen participating teams manually annotated thirty evidence samples each, five of which were gold-labeled samples annotated by ourselves.
For the final set, we only considered annotation submissions of a team, if the gold-labeled samples were correctly annotated, resulting in a final set consisting of $278$ evidence annotations.
To measure the agreement between annotators, we used two methods. First, we calculated the agreement between the categorical verdict labels that annotators selected as the first step in the annotation process. Therefore, we used all $278$ annotated samples after filtering as specified above, and obtained a Krippendorff's Alpha (K-$\alpha$)~\citep{krippendorff2004content} $0.732$ and a Fleiss' Kappa~\citep{fleiss1971measuring} (F-$\kappa$) of $0.727$ - both indicating substantial agreement. Second, for the rating dimensions which are annotated with numeric values between one and five, we calculated the standard deviation (std) among the rating, resulting in std between $0.804$ and $1.318$ (see \Cref{table:annotator_agreement}).

\begin{table}[ht]
    \centering
    \scalebox{0.9}{
    \begin{tabular}{
    p{0.4\linewidth} p{0.38\linewidth} p{0.15\linewidth}} \toprule
        Rating Dimension & IAA method & Score \\\midrule
        Verdict agreement & Fleiss-$\kappa$ & 0.727 \\
        Verdict agreement & Krippendorff-$\alpha$ & 0.732 \\
        Coverage & Std & 1.119 \\
        % Coherence & Std & 1.013 \\
        % Redundancy & Std & 1.317 \\
        % Consistency & Std & 0.921 \\
        Relevance & Std & 0.804 \\
    \bottomrule
    \end{tabular}}
    \caption{Inter-annotator agreement (IAA) scores for human ratings using Fleiss-$\kappa$/Krippendorff-$\alpha$ for categorical (i.e., verdict agreement) and standard deviation (Std) for continuous ratings.}
    \label{table:annotator_agreement}
    % \vspace{-1em}
\end{table}

\paragraph{Evaluation dimensions}

With the help of human raters, we assessed how well the scorers correlate with human judgments. For this evaluation, we considered three dimensions: 

\noindent \textbf{$(1)$ Coverage} assesses how much of the reference evidence is covered by the retrieved evidence, whether the content, meaning, entities, etc., of the reference evidence are fully represented in the retrieved evidence, ensuring that the retrieved evidence covers all aspects of the reference evidence necessary for the evaluation. Partial evidence may overlook certain details and result in incomplete or incorrect conclusions.

\noindent \textbf{$(2)$ Relevance} measures the relevance of the evidence retrieved for the claim, ensuring that the evaluated evidence addresses the claim and that minimal unrelated information is included.

\noindent \textbf{$(3)$ Verdict Agreement} measures if the retrieved evidence results in the same verdict label as the reference evidence. Discrepancies in verdicts can indicate issues with the retrieved evidence such that a follow-up review can be performed to ensure accuracy in the fact-checking process.

While coverage can penalise potentially valid evidence, this only occurs when such evidence is included at the expense of the gold/reference evidence. As long as the gold evidence is fully covered, the coverage score remains high, even if additional content is included. This differs from the relevance score we introduced above, which is more flexible: it does not penalise the inclusion of content different to the gold evidence, as long as the content is relevant for verifying the claim. We report both alignment with relevance and coverage, with the latter being relevant for the recall-oriented automatic Ev\textsuperscript{2}R evaluation.

\subsection{Evaluation with additional benchmarks}
\label{ssec:evaluation_afc_benchmarks}

In addition to the AVeriTeC shared task data, we use the large-scale AFC benchmarks FEVER~\citep{thorne-etal-2018-fever} and VitaminC~\citep{schuster-etal-2021-get}. 
FEVER is a benchmark designed for automated claim verification against textual sources from Wikipedia. It contains $185.4k$ claims generated through modifying sentences from Wikipedia and classified as supported/refuted/not enough information.
VitaminC is an AFC benchmark designed for robust claim verification. It consists of approximately $400k$ pairs of claims and evidence pairs derived from Wikipedia revisions. The aim was to create contrastive pairs of nearly identical evidence, differing only by subtle factual details, such that one evidence set supports the claim and the other refutes it, or vice versa.
For scorer assessment, we created tuples consisting of a claim $c_i$, a reference evidence set $E_{ref}$, predicted evidence set $E_{pred}$ from the multiple evidence sets available for individual claims within the FEVER and VitaminC test datasets. If both evidence sets resulted in the same verdict label, we assigned a verdict agreement score of $1$; otherwise, we assigned a score of $0$. The resulting data were used for evaluations presented in \Cref{tab:results_verdict_agree}.

\subsection{Baselines}
\label{ssec:baselines}

To validate the effectiveness of Ev\textsuperscript{2}R, we compare it to three types of baseline scorers: $(i)$ the reference-based BLEURT scorer, which compares retrieved evidence to reference evidence; $(ii)$ proxy-reference scorers, which assesses evidence through predicted verdict labels; and $(iii)$ reference-less scorers, which rely only on the claim and retrieved evidence.\footnote{Prompts and further details are given in the appendix.}

\paragraph{$(i)$ Reference-based baseline scorers}

As a reference-based baseline, we adapt a scorer built on BLEURT~\citep{sellam-etal-2020-bleurt}, originally developed for NLG tasks such as summarization. BLEURT uses BERT as its backbone and is trained on millions of synthetically generated sentence pairs to capture semantic similarities between texts.
To evaluate the retrieved evidence against reference evidence, the BLEURT-based scorer compares both evidence sets and gives a similarity score. The final aggregated score $s$ is calculated by averaging instance-level scores across all evaluated samples.

Given a set of retrieved evidence $\hat{E} = \{\hat{e}_1, \hat{e}_2, \ldots, \hat{e}_n\}$ and reference evidence $E = \{e_1, e_2, \ldots, e_n\}$, the scorer computes:
\[
s = \text{BLEURT}(E, \hat{E})
\]
where $s \in [0, 1]$, with higher values indicating greater semantic similarity between the evidence.

% We fine-tune BLEURT for evidence evaluation using synthetically generated input pairs of the form $(c_i, e_i)$ and $(c_i, \hat{e}_i)$, where $c_i$ is a claim, $e_i$ is reference evidence, and $\hat{e}_i$ is retrieved evidence. We extract training data from large-scale benchmarks widely used in AFC, specifically FEVER~\citep{thorne-etal-2018-fever}, VitaminC~\citep{schuster-etal-2021-get}, and Hover~\citep{jiang-etal-2020-hover}, which provide multiple evidence sets ${E_1, E_2, \ldots, E_n}$ per claim.

We fine-tune BLEURT for evidence evaluation using the Pytorch version of the original BLEURT model\footnote{\href{https://huggingface.co/Elron/bleurt-base-512}{Elron/bleurt-base-512}} and the HuggingFace Trainer API over 5 epochs, with a batch size of 4 and learning rate set to $1\text{e}^{-5}$. The model is trained on claim-evidence input pairs from AFC datasets including FEVER~\citep{thorne-etal-2018-fever}, VitaminC~\citep{schuster-etal-2021-get}, and HOVER~\citep{jiang-etal-2020-hover}. These datasets provide multiple annotated evidence sets ${E_1, E_2, \ldots, E_n}$ per claim, enabling the construction of diverse similarity pairs. we generate input-label pairs where positive examples (similarity score $s = 1$) are created using two distinct evidence sets aligned to the same claim or based on matching claim-verdict pairs (e.g., both supporting the claim). Negative examples ($s = 0$) are created by pairing evidence from different claims or by contrasting verdict-aligned pairs (e.g., one supporting, one refuting).

% 
% For positive examples ($s = 1$), we use two distinct evidence sets that are annotated for the same claim. For negative examples ($s = 0$), we pair evidence from different claims (e.g., $c_i$ and $c_j$). Moreover, we use the label diversity in the VitaminC dataset by considering evidence sets with matching labels (e.g., both supporting the claim) as positive evidence pairs and those with conflicting labels (e.g., one supporting and one refuting the claim) as negative ones.

Additionally, we include metrics such as RougeL, BLEU, METEOR, and Hungarian METEOR as reference-based baselines since they have been previously also used in AFC for evidence evaluation~\citep{schlichtkrull2024automatedverificationtextualclaims}. They were originally developed for NLG tasks and compare retrieved ($\hat{E}$) with reference evidence ($E$) based on surface level or semantic similarity.

\paragraph{$(ii)$ Proxy-reference LLM baseline}
As a proxy-reference baseline, we propose an LLM-based scorer that estimates the likelihood of the correct veracity label $y$ given the claim $c$ and the retrieved evidence $\hat{E}$. Instead of relying on a predicted probability distribution as in trained classifiers, we prompt the LLM directly with an instruction to predict the veracity label and use its log-probability for the correct label as the proxy score. The scorer computes a score $s$ as follows:
\[
s = \log p(y \mid P(I, c, \hat{E}), \theta)
\]
where $P(I, c, \hat{E})$ is the prompt including an instruction $I$, the input claim $c$, and the retrieved evidence $\hat{E}$, and $\theta$ denotes the parameters of the LLM.
We apply chain-of-thought prompting~\citep{Wei0SBIXCLZ22} and instruct the model to generate intermediate reasoning steps before selecting a final label. The prompt includes few-shot examples and reasoning steps as demonstrations to improve prediction quality.
For the proxy-reference scorer, we use DeBERTa as the backbone model. Our choice is motivated by its strong performance on claim verification and natural language inference (NLI) tasks under fine-tuned settings, as well as its compatibility with our available compute resources (one A$100$ GPU).

\paragraph{$(iii)$ Reference-less baseline}
As a reference-less baseline, we instruct an LLM-based scorer evaluate the retrieved evidence $\hat{E}$ based solely on the input claim $c$, without relying on any reference evidence. The scorer decomposes the claim into atomic facts $A_c$ and checks whether each fact $a_c \in A_c$ is addressed, for example, supported or refuted, by the retrieved evidence. The score is computed as follows:
\[
s = \frac{1}{|A_c|} \sum_{a_c \in A_c} \mathbb{I}[a_c \text{ supported/refuted by } \hat{E}]
\]
% Whereas, the indicator function $\mathbb{I}$ returns 1 if $a_c$ is supported or refuted by $\hat{E}$, and 0 otherwise.

This approach, similar to the Ev\textsuperscript{2}R reference-based component, is inspired by FactScore~\citep{min-etal-2023-factscore}. However, unlike the reference-based evaluation, which extracts evidence for both gold and retrieved evidence, this method decomposes the claim to evaluate the retrieved evidence without relying on any additional reference.\footnote{The reference-less evaluation prompt is given in \Cref{fig:prompt_reference_less_component} in the appendix.} 

\paragraph{Backbone models}

For the LLM-based scorers, we evaluate several backbone models: GPT$4o$, Gemini $1.5$ Pro, Gemini $1.5$ Flash, and Llama 3.1 $70$B (see \Cref{tab:results_verdict_agree}). 
GPT$4o$ is a state-of-the-art model that demonstrates strong performance across a wide range of benchmarks and domains~\citep{GPT4o-introduction}. We also evaluate Gemini $1.5$ Pro, which has shown strong performance in reasoning and long-context understanding tasks~\citep{Gemini1.5}, while Gemini $1.5$ Flash, a smaller variant, allows us to explore scalability across different model sizes. We also include Llama $3.1$ $70$B~\citep{Dubey-etal-2024} to understand how well our proposed scorers perform with state-of-the-art open source models compared to the previously mentioned closed-source models.

\section{Results and Discussion}

In this section, we discuss the results obtained with Ev\textsuperscript{2}R using three different data sources. First, human-rated system predictions from the AVeriTeC shared task (see~\Cref{tab:results_averitec_shared_task_trimmed}). Second, the FEVER and VitaminC test sets for which we obtained pairs of evidence following the approach outlined in \Cref{ssec:baselines}. Third, adversarial tests as described in ~\Cref{tab:results_checklist}.
Following \citet{FuNJL23}, we measure the correlation between the scorers and human ratings using Spearman ($\rho$)~\citep{Spearman1987} and Pearson correlation coefficients ($r$)~\citep{Pearson1896}.

\begin{table*}[!ht]
\centering
\scalebox{0.8}{
\begin{tabular}{l | c c | c c | c c | c c }
\hline
\textbf{Scorer} & \multicolumn{2}{c|}{\bf VitaminC} & \multicolumn{2}{c|}{\bf FEVER} & \multicolumn{2}{c|}{\bf AVeriTeC} & \multicolumn{2}{c}{\bf Avg} \\
 & $\rho$ & $r$ & $\rho$ & $r$ & $\rho$ & $r$ &  $|\rho|$ & $|r|$ \\
 \hline
\multicolumn{9}{l}{ \textbf{Reference-based Baselines}} \\[0.2em]
% \hline
BLEURT & -.065 & -.072 & -.003 & -.000 & .151 & .098 & .073 & .057 \\
RougeL & .030 & .022 & -.010 & -.009 & .070 & .097 & .037 & .043 \\
BLEU & .030 & .026 & -.006 & -.003 & .166 & .137 & .067 & .055 \\
Meteor & .030 & .021 & -.010 & -.003 & .152 & .159 & .064 & .061 \\
H-METEOR & .001 & -.001 & -.027 & -.016 & -.029 & -.019 & .019 & .012 \\
 \hline
\multicolumn{9}{l}{ \textbf{Reference-less Baselines}} \\[0.2em]
GPT$4o$ & .014 & .015 & \textcolor{skyblue}{.457} & \textcolor{skyblue}{.468} & .240 & .243 & .237 & .242 \\
Gemini-Pro & .034 & .033 & .415 & .412 & .262 & .273 & .237 & .239 \\
Gemini-Flash & .038 & .045 & \textcolor{orangebrown}{.434} & \textcolor{orangebrown}{.426} & .230 & .247 & .234 & .239 \\
Llama 3.1 & .024 & .034 & .293 & .308 & .215 & .208 & .177 & .183 \\
\hline
\multicolumn{9}{l}{\textbf{Ev\textsuperscript{2}R Weighted Scorer}} \\[0.2em]
GPT-4o / Proxy-ref & .276 & .318 & .366 & .360 & \textcolor{skyblue}{.500} & \textcolor{skyblue}{.488} & \textcolor{skyblue}{.381} & \textcolor{skyblue}{.389} \\
Gemini-Pro / Proxy-ref & \textcolor{orangebrown}{.295} & \textcolor{orangebrown}{.334} & .349 & .346 & \textcolor{orangebrown}{.472} & \textcolor{orangebrown}{.464} & \textcolor{orangebrown}{.372} & \textcolor{orangebrown}{.381} \\
Gemini-Flash / Proxy-ref & \textcolor{skyblue}{.336} & \textcolor{skyblue}{.346} & .170 & .240 & .129 & .197 & .292 & .294 \\
Llama / Proxy-ref & .272 & .322 & .026 & .227 & .324 & .307 & .243 & .273 \\
\hline
\hline
\end{tabular}}
\caption{\label{tab:results_verdict_agree} Correlation on the \textbf{Verdict Agree} category calculated using the Spearman ($\rho$) and Pearson ($r$) correlation coefficients for the datasets VitaminC, FEVER, and AVeriTeC (shared task subset). Averages are computed over the absolute values of each metric per row. \textcolor{skyblue}{Highest scores per column} are colored blue and \textcolor{orangebrown}{second-highest values} highlighted with brown color. 
% \textcolor{darkgray}{Reference-based precision and recall} scores which are calculated to compute the final reference-based $F_1$ score, are colored in gray. 
}
% \vspace{-1em}
\end{table*}

\begin{table*}[!ht]
\centering
\scalebox{0.8}{
\begin{tabular}{l | c c | c c | c c | c c }
\hline
\textbf{Scorer} & \multicolumn{2}{c|}{\bf VitaminC} & \multicolumn{2}{c|}{\bf FEVER} & \multicolumn{2}{c|}{\bf AVeriTeC} & \multicolumn{2}{c}{\bf Avg} \\
 & $\rho$ & $r$ & $\rho$ & $r$ & $\rho$ & $r$ &  $|\rho|$ & $|r|$ \\
\hline
\multicolumn{9}{l}{ \textbf{Reference-based} (Precision)} \\[0.2em]
GPT$4o$ & .203 & .200 & \bf.140 & \bf.129 & \bf.278 & \bf.256 & \bf.207 & \bf.195 \\
Gemini-Pro & \bf.222 & \bf.216 & .139 & .118 & .237 & .177 & .199 & .170 \\
Gemini-Flash & .202 & .198 & .106 & .079 & -.290 & -.205 & .199 & .161 \\
Llama 3.1 & -.037 & -.036 & .110 & .109 & .216 & .210 & .121 & .118 \\
\hline
\multicolumn{9}{l}{ \textbf{Reference-based} (Recall)} \\[0.2em]
GPT$4o$ & .229 & .228 & .146 & .114 & \bf.314 & \bf.284 & .230 & .209 \\
Gemini-Pro & \bf.253 & \bf.246 & \bf.176 & \bf.143 & .285 & .278 & \bf.238 & \bf.222 \\
Gemini-Flash & .232 & .227 & .032 & .000 & -.190 & -.039 & .151 & .089 \\
Llama 3.1 & .020 & .025 & -.011 & -.012 & .128 & .121 & .053 & .053 \\
\hline
\multicolumn{9}{l}{ \textbf{Reference-based} ($F_1$)} \\[0.2em]
% \multicolumn{9}{l}{ Prompt Scorer GPT$4o$} \\[0.2em]
% \hline
% \textcolor{darkgray}{Ref-based (prec)} & \textcolor{darkgray}{.203} & \textcolor{darkgray}{.200} & \textcolor{darkgray}{.140} & \textcolor{darkgray}{.129} & \textcolor{darkgray}{.278} & \textcolor{darkgray}{.256} & \textcolor{darkgray}{.207} & \textcolor{darkgray}{.195} \\
% \textcolor{darkgray}{Ref-based (recall)} & \textcolor{darkgray}{.229} & \textcolor{darkgray}{.228} & \textcolor{darkgray}{.146} & \textcolor{darkgray}{.114} & \textcolor{darkgray}{.314} & \textcolor{darkgray}{.284} & \textcolor{darkgray}{.230} & \textcolor{darkgray}{.209} \\
GPT$4o$ & .215 & .213 & .143 & .121 & \bf.295 & \bf.269 & \bf.218 & \bf.202 \\
Gemini-Pro & \bf.237 & .\bf230 & \bf.155 & \bf.129 & .259 & .216 & .217 & .193 \\
Gemini-Flash & .216 & .212 & .049 & .000 & -.230 & -.066 & .172 & .115 \\
Llama 3.1  & .087 & .164 & -.024 & -.027 & .161 & .154 & .074 & .073\\
% Proxy-ref & .031 & .034 & .391 & .482 & .523 & .483 & .315 & .333 \\
\hline
\multicolumn{9}{l}{ \textbf{Proxy-reference}} \\[0.2em]
% \hline
Proxy-ref & .456 & .479 & .290 & .480 & .487 & .459 & .411 & .473 \\
\hline
\hline
\end{tabular}}
\caption{\label{tab:results_verdict_agree_components_details} More detailed scores for both Ev\textsuperscript{2}R components: the reference-based part (using different LLMs as backbone models) and the DeBERTa-based proxy-reference component.} 
\end{table*}

\paragraph{General insights across datasets}

Overall, the trained proxy-reference scorer yields the highest overall correlation across the datasets we assess, except for AVeriTeC (see \Cref{tab:results_verdict_agree}). Unlike the other benchmarks, which contain claims and evidence extracted from Wikipedia~\citep{thorne-etal-2018-fever, schuster-etal-2021-get}, AVeriTeC contains more complex, real-world claims extracted from fact-checking websites and evidence from the web. Hence, the trained proxy-reference scorer is outperformed by the more recent LLM GPT$4o$
on AVeriTeC.
Moreover, across almost all datasets in \Cref{tab:results_verdict_agree} and all rating categories in \Cref{tab:results_averitec_shared_task_trimmed}, our Ev\textsuperscript{2}R scorers rank among the top-$2$ scorers.
The obtained insights support the practical usefulness of the Ev\textsuperscript{2}R scorer for evaluating evidence in realistic AFC scenarios.

\begin{table*}[!h]
\centering
\scalebox{0.66}{
\begin{tabular}{l | c c c c | c | c c c c | c c c c}
\hline
 & \multicolumn{4}{c|}{\bf Ref-less Baselines} & \textbf{Proxy-ref} & \multicolumn{4}{c|}{\bf Ref-based Baselines} & \multicolumn{4}{c}{\bf Ev\textsuperscript{2}R} \\
 & GPT-4o & Gem-Pro & Gem-Flash & Llama & DeBERTa & RougeL & BLEU & Meteor & H-Met & GPT-4o & Gem-Pro & Gem-Flash & Llama \\
\hline
COV ($\rho$) & .237 & .287 & .275 & .297  & \textcolor{orangebrown}{.338} & .150 & .236 & .229 & .005 & .321 & \textcolor{skyblue}{.341} & .221 & .328 \\
COV ($r$) & .261 & .296 & .287 & .286 & \textcolor{skyblue}{.348} & .169 & .184 & .240 & -.024 & .323 & \textcolor{orangebrown}{.326} & .299 & .336 \\
REL ($\rho$) & .292 & .203 & .287 & .227  & \textcolor{orangebrown}{.298} & .086 & .107 & .062 & .008 & .297 & .278 & \textcolor{skyblue}{.403} & .224 \\
REL ($r$) & .360 & .263 & .333 & .250 & \textcolor{orangebrown}{.374} & .099 & .079 & .076 & .003 & .332 & .315 & \textcolor{skyblue}{.404} & .277 \\
\hline
\end{tabular}}
\caption{\label{tab:results_averitec_shared_task_trimmed} Correlation between human-rated \textbf{AVeriTeC shared task submissions} and proposed scorers for COVerage and RELevance dimensions. \textcolor{skyblue}{Highest scores per dimension} are colored blue and \textcolor{orangebrown}{second-highest values} highlighted with brown color.}
\end{table*}

Further analysis across all datasets demonstrates that the evaluated Ev\textsuperscript{2}R scorer components result in statistically significant correlations across backbone models selected for reference-based scoring. 
We assessed statistical significance using two-sided hypothesis tests on the correlation coefficients, as implemented in \texttt{scipy.stats.pearsonr} and \texttt{scipy.stats.spearmanr}, with the null hypothesis that no correlation (i.e., $r = 0$ or $\rho = 0$) is present. We conducted all tests using $n = 278$ data points from the AVeriTeC dataset for which we had collected human ratings as specified in \Cref{ssec:human_eval}, as well as the FEVER and VitaminC test sets preprocessed as specified in \Cref{ssec:evaluation_afc_benchmarks}.
The proxy reference scoring component achieves the strongest correlations across datasets ($\rho=0.375$, $r=0.393$ for FEVER; $\rho=0.250$, $r=0.273$ for VitaminC; $\rho=0.475$, $r=0.450$ for AVeriTeC; all $p\text{-values}<0.001$). Among the other models, GPT-4o and Gemini-Pro generally demonstrate good predictive power (with all $p\text{-values}<0.001$), while Gemini-Flash results in weaker, but still statistically significant correlations. Our results show variability in models for reference scoring across benchmarks.\footnote{See \Cref{table:results_significance} for detailed correlation results.}

% How do the proposed prompt scorers compare against traditional metrics in terms of correlation with human ratings?
\paragraph{Ev\textsuperscript{2}R weighted score versus traditional metrics}

Across all datasets, i.e., VitaminC, FEVER, and AVeriTeC, the Ev\textsuperscript{2}R scorer consistently outperforms traditional reference-based baselines we assess, including ROUGE-L, BLEU, METEOR, and H-METEOR. The performance gap is particularly prevalent on AVeriTeC, which includes real-world claims and evidence retrieved from the web. This setting is notably more complex than earlier benchmarks like FEVER and VitaminC. In such cases, traditional metrics that rely on surface-level or shallow semantic similarity often fail to generalize. 
These findings are supported by the insights we gain from the adversarial tests in \Cref{tab:results_checklist}, where we assess all baselines against both both Ev\textsuperscript{2}R components, the proxy reference (last column) and reference-based (prec/recall) components. On all semantics-preserving tests, both Ev\textsuperscript{2}R components yield robustness performance. 

\paragraph{Ev\textsuperscript{2}R: Balancing proxy and reference-based evidence evaluation}
While the proxy-reference component of Ev\textsuperscript{2}R helps stabilize scores on academic datasets with short claims and evidence (e.g., FEVER, VitaminC in \Cref{tab:results_verdict_agree}), relying solely on proxy signals proves insufficient for more complex real-world scenarios such as AVeriTeC (see \Cref{tab:results_verdict_agree}).
Prior work~\citep{mccoy-etal-2019-right} has shown that NLI models often exploit shallow syntactic heuristics such as lexical overlap or subsequence matching that can lead to correct predictions for the wrong reasons. This motivated the need for a balanced evaluation approach that combines proxy-reference with reference-based assessment and offers a more reliable and interpretable measure of evidence data.
Moreover, for relevant assessment dimensions such as \textit{Coverage} of reference evidence and \textit{Relevance} of retrieved evidence, the reference-based component of Ev\textsuperscript{2}R yields higher correlation with human assessment.

\begin{table*}[!htbp]
\centering
\scalebox{0.75}{
\begin{tabular}{l c c c c c | c c c | c }
\hline
\textbf{Dataset} &  \multicolumn{5}{c|}{\bf Reference-less Baselines} & \multicolumn{3}{c|}{\bf Ev\textsuperscript{2}R} & \bf Proxy-ref  \\
 & METEOR & ROUGE & BLEU & BLEURT & GPT4o & Prec & Rec & Proxy-comp. & DeBERTa \\
\hline
\hline
\multicolumn{10}{l}{\bf Semantics-altering tests}\\
Completeness & -30.56 & -53.19 & -77.87 & \textbf{\textcolor{skyblue}{-143.70}} & \textbf{\textcolor{orangebrown}{-91.76}} & -21.2 & -59.5 & -39.35 & -82.68 \\
Random shuffle & -32.32 & -89.94 & \textbf{\textcolor{skyblue}{-94.20}} & -1.02 & 0.55 & -6.8 & -6.0 & -32.22 & \textbf{\textcolor{orangebrown}{-92.32}} \\
\hline
\textbf{Average} & -31.44 & -71.57 & \textbf{\textcolor{orangebrown}{-86.04}} & -72.36 & -45.61 & -14.0 & -32.75 & -35.79 & \bf \textcolor{skyblue}{-87.50} \\
\hline
\hline
\multicolumn{10}{l}{\bf Semantics-preserving tests}\\
Invariance contraction & -0.53 & -0.97 & -1.35 & -1.65 & \textbf{\textcolor{skyblue}{0}} & -0.8 & -0.7 & -0.85 & \textbf{\textcolor{orangebrown}{0.02}} \\
Invariance num2text & -8.34 & -10.24 & -14.10 & -22.13 & -3.85 & \textbf{\textcolor{orangebrown}{0.1}} & \textbf{\textcolor{skyblue}{0}} & -4.02 & -35.64 \\
Invariance text2num & -0.04 & -0.90 & -1.15 & -0.81 & -4.40 & \bf \textcolor{skyblue}{0} & -0.3 & -1.59 & \bf \textcolor{orangebrown}{0.01} \\
Invariance synonyms & \textbf{\textcolor{orangebrown}{-8.22}} & -36.64 & -51.83 & -51.00 & \textbf{\textcolor{skyblue}{-2.75}} & -11.6 & -11.3 & -15.35 & -85.71 \\
Redundancy sent & -26.16 & -46.52 & -56.15 & -80.45 & 3.30 & \bf \textcolor{skyblue}{0} & \bf \textcolor{orangebrown}{-0.1} & -5.04 & -67.95 \\
Redundancy words & -5.73 & -10.68 & -32.69 & -31.90 & -2.75 & \bf \textcolor{orangebrown}{-1.0} & \bf \textcolor{skyblue}{0.2} & -2.89 & -52.55 \\
Fluency & -11.52 & -23.33 & -29.26 & -41.69 & \bf \textcolor{skyblue}{-2.20} & -4.3 & -3.9 & \bf \textcolor{orangebrown}{-2.6} & -57.12 \\
Noise & -31.82 & -31.81 & -31.81 & -21.20 & \bf \textcolor{orangebrown}{-1.65} & -28.7 & -2.2 & \bf \textcolor{skyblue}{2.89} & -62.42 \\
Argument structure & -0.01 & \bf \textcolor{orangebrown}{0} & \bf \textcolor{orangebrown}{0} & \bf \textcolor{orangebrown}{0} & \bf \textcolor{skyblue}{1.65} & \bf \textcolor{orangebrown}{0} & \bf \textcolor{orangebrown}{0} & -0.74 & -28.93 \\
\hline
\textbf{Average} & -10.33 & -18.43 & -23.94 & -27.87 & -15.04 & -5.16 & \bf \textcolor{skyblue}{-2.9} & \bf \textcolor{orangebrown}{-3.78} & -39.79 \\
\hline
\hline
\end{tabular}}
\caption{\label{tab:results_checklist} Results obtained for adversarial tests as score difference (in \%) between initial evidence and manipulated evidence. The evaluated Ev\textsuperscript{2}R reference-based scoring uses $GPT4o$. Overall, the reference-based scoring and the proxy-reference scoring components demonstrate the most robust performance on adversarial tests, together with the trained reference-based baselines, as evidenced by small performance drops on semantics-preserving tests.
\textcolor{skyblue}{Best scores per evaluation category} are colored blue and \textcolor{orangebrown}{second-best values} highlighted with brown color. }
\end{table*}

% How does the selection of backbone models influence the prompt scorer's performance?
\paragraph{Selection of backbone LLMs}

Assessing the impact of different backbone LLMs for the reference-based component of Ev\textsuperscript{2}R, we find that the model selection has only a limited effect on the overall performance of the Ev\textsuperscript{2}R scorer across datasets (\Cref{tab:results_verdict_agree}) and evaluation dimensions (\Cref{tab:results_verdict_agree} and \ref{tab:results_averitec_shared_task_trimmed}).
No single model consistently outperforms others across all dimensions and datasets. While scorer variants using Gemini models correlate higher with human assessments in the dimensions coverage and relevance, the GPT-$4o$-based variant excels in the category of verdict agreement (\Cref{tab:results_verdict_agree}). Moreover, the combination of GPT-$4o$ and proxy-reference shows strong correlations on AVeriTeC and is competitive overall.

% No proxy-reference scorer for the Gemini-based scorer as the model does not provide the log probability of predictions.

% \item How are invariance tests (e.g., contraction, synonyms) affecting the scoring outcomes, and what insights can we gain about the robustness of the different models?
% \item How does the presence of noise and redundancy in the data affect the performance of different scorers and what insights can we gain regarding scorers' limitations?

\paragraph{Adversarial tests}
In \Cref{tab:results_checklist}, we assess the performance of Ev\textsuperscript{2}R components (reference-based and proxy-reference) across various adversarial tests against baselines, considering both semantics-preserving and semantics-altering tests.
Our insights highlight how Ev\textsuperscript{2}R components (i.e., reference-based and proxy-reference) versus baseline scorers respond to adversarial manipulations in evidence texts. Traditional metrics such as METEOR, ROUGE, BLEU, and especially BLEURT show high sensitivity to semantics-altering tests, demonstrated through substantial performance drops. On the other hand, the Ev\textsuperscript{2}R components, specifically proxy-reference scoring, exhibit stronger robustness to both semantics-altering and semantics-preserving evaluation. These findings highlight their suitability and robustness for evidence evaluation. Overall, Ev\textsuperscript{2}R demonstrates greater robustness compared to baselines in adversarial settings. 

\subsection{Qualitative analysis of evidence evaluation}

% Through manual assessment of evaluated evidence, we observe various strengths but also some limitations of the proposed Ev\textsuperscript{2}R scorer.

\paragraph{Reference-based}

Reference evidence provides additional context to the Ev\textsuperscript{2}R scorer, which can be leveraged during the evaluation of retrieved evidence.
The reference-based scoring component of Ev\textsuperscript{2}R allows a more detailed and interpretable evaluation by breaking down both reference and predicted evidence into atomic facts (see \Cref{table:ref_based_evidence_eval_examples} in the appendix).  Reporting separate precision and recall scores facilitates a more detailed understanding of the retrieved evidence. For instance, retrieved evidence sometimes contains excessive details compared to the reference, negatively affecting precision but still yielding high recall.

While this approach enhances the evaluation process,  our qualitative analysis identified challenges that are addressed by stabilizing the Ev\textsuperscript{2}R scorer through integration of a proxy-reference component.
One specific challenge for reference-based scoring are cases where predicted evidence uses different information or follows a different reasoning chain than the reference evidence, but results in the same verdict (e.g., refuting a claim, as illustrated in \Cref{table:ref_based_evidence_eval_examples} in the appendix). In such cases, the reference-based scoring tends to yield lower scores compared to reference-less and proxy-reference scorers. For example, in \Cref{table:ref_based_evidence_eval_examples} both reference and predicted evidence contradict the claim that Nigeria is the fifth largest recipient of diaspora remittances. However, the predicted evidence states that Nigeria is the sixth largest recipient, while the reference states it is the seventh largest. Hence, the scorer identifies a mismatch between the reference and predicted evidence. Furthermore, retrieved evidence can sometimes include more details than the reference, which impacts precision but leads to a high recall. To address these limitations and stabilise the Ev\textsuperscript{2}R scorer, we incorporate proxy-reference scoring.

\paragraph{Proxy reference}

We also observe that a key challenge for proxy-reference scoring are evidence pairs where both predicted and reference evidence are related but result in diverging verdict labels (see \Cref{table:proxy_reference_evidence_eval_examples} in the appendix). 
For example, considering the claim \textit{``Scotland is spending more on health per head than the rest of the UK.''} in \Cref{table:proxy_reference_evidence_eval_examples}; while the predicted evidence closely resembles the reference in surface-level characteristics, the predicted evidence supports the claim, whereas the reference evidence does not, resulting in a low proxy-reference score.
Similarly, minor numerical differences, such as those observed in evidence in \Cref{table:proxy_reference_evidence_eval_examples}, can also lead to diverging verdicts despite overall consistency between the reference and predicted evidence.
In these cases, a more interpretable assessment based on precision and recall scores from reference-based scoring can yield further insights and enhance readers' understanding of the presented evidence.

\subsection{How \textcolor{skyblue}{Ev\textsuperscript{2}}\textcolor{orangebrown}{R} addresses key challenges of evidence evaluation for AFC}

This section reiterates key challenges of evidence evaluation (see Sections \ref{sec:introduction} and \ref{sec:related_work}) and Ev\textsuperscript{2}R.
% and discusses how the proposed Ev\textsuperscript{2}R scorer addresses them.

\paragraph{Implicit evidence assessment}

Verdict-based evaluation approaches can inaccurately rate evidence high by predicting correct verdicts without genuinely relying on the retrieved evidence. To overcome this over-reliance on verdict alone, Ev\textsuperscript{2}R explicitly evaluates both alignment with reference evidence and verdict agreement. Evaluation results from AVeriTeC, FEVER, and VitaminC show high correlation with human judgments on criteria such as verdict agreement, evidence relevance, and coverage of reference data, indicating Ev\textsuperscript{2}R assesses retrieved evidence rather than verdict correctness alone.

\paragraph{Assessment relying on closed knowledge sources \& surface-level matching of evidence}

Previous widely-used approaches for evidence evaluation (e.g., the FEVER score~\citep{thorne-etal-2018-fever}) relied on predefined sources (e.g., Wikipedia) and exact matches between predicted and reference evidence. Such methods are not extendable to realistic AFC settings, where predicted evidence may not be part of the annotated corpus. Additionally, token-matching metrics (e.g., h-METEOR used in \citep{schlichtkrull2024automatedverificationtextualclaims}) are overly sensitive to surface-level differences and fail to capture alternative valid evidence or reasoning paths. Ev\textsuperscript{2}R's reference-based component decomposes retrieved evidence into atomic facts and assesses their alignment with reference facts beyond strict matching constraints. Evaluations of evidence predicted with shared task systems (see \Cref{tab:results_averitec_shared_task_trimmed}), which include unannotated web evidence, confirm the effectiveness of this approach. Furthermore, adversarial test results (see \Cref{tab:results_checklist}) demonstrate Ev\textsuperscript{2}R's superior robustness against surface-level perturbations compared to baseline scorers.

\paragraph{Multiple valid evidence sets and reasoning paths \& temporal constraints with evidence annotation}

For a given claim, multiple correct sets of evidence or alternative reasoning paths can exist. Previous evaluation methods might penalize valid alternative evidence. Ev\textsuperscript{2}R with its both scoring compnents considers scenarios with multiple valid reasoning paths and evidence absent in the annotated references. Evaluations using the datasets FEVER, VitaminC, and AVeriTeC, where exact matching between reference and predicted evidence is not expected, show Ev\textsuperscript{2}R achieves higher correlations with human judgments and stable performance under adversarial testing.

\section{Conclusion}

This paper introduces Ev\textsuperscript{2}R, a weighted evidence evaluation scorer for AFC. Ev\textsuperscript{2}R consists of a reference-based and a verdict-level proxy evaluation component and addresses limitations in previous evaluation approaches by jointly assessing how well evidence aligns with gold references and how reliably the evidence supports the verdict. We assessed Ev\textsuperscript{2}R against three baseline approaches for evidence evaluation approaches and find that Ev\textsuperscript{2}R correlates strongly with human judgments on relevant evaluation criteria and is a robust scorer for evidence evaluation in AFC.

\section{Limitations \& Ethics statement}

This work has several limitations that are important to acknowledge. First, the datasets used are restricted to English, limiting the generalisability of findings to global fact-checking contexts where multilingual capabilities are essential. This English-only focus reflects a broader issue in the field, where most benchmarks are available solely in English, potentially distorting perceptions of progress in automated fact-checking. Additionally, all labels, e.g., supports and refutes used in this work to classify claims, capture the relationship between claims and evidence, they do not imply any real-world correctness of the claims. Finally, We informed the participants about the data being collected and its purpose. Participants had the opportunity to withdraw at any time and to provide feedback.

% \section{Limitations}

% \todo{Discuss why we don't test \textit{trained reference-less} scorers. What would be needed to achieve this in future research?}

 \section*{Acknowledgements}
 We thank Nedjma Ousidhoum, Julian Eisenschlos, Oana Cocarascu, and Elena Simperl for their valuable feedback and suggestions about this work. Mubashara Akhtar acknowledges funding from the ETH AI Center Postdoctoral Fellowship. Andreas Vlachos is supported by the ERC grant AVeriTeC (GA 865958). Michael Schlichtkrull is supported by the EPSRC grant AdSolve (EP/Y009800/1), a RAI UK Keystone project.

\bibliography{tacl2021, anthology, references}

\begin{thebibliography}{55}
\expandafter\ifx\csname natexlab\endcsname\relax\def\natexlab#1{#1}\fi

\bibitem[{Akhtar et~al.(2023)Akhtar, Schlichtkrull, Guo, Cocarascu, Simperl, and Vlachos}]{akhtar-etal-2023-multimodal}
Mubashara Akhtar, Michael Schlichtkrull, Zhijiang Guo, Oana Cocarascu, Elena Simperl, and Andreas Vlachos. 2023.
\newblock \href {https://doi.org/10.18653/v1/2023.findings-emnlp.361} {Multimodal automated fact-checking: A survey}.
\newblock In \emph{Findings of the Association for Computational Linguistics: EMNLP 2023}, pages 5430--5448, Singapore. Association for Computational Linguistics.

\bibitem[{Akhtar et~al.(2024)Akhtar, Subedi, Gupta, Tahmasebi, Cocarascu, and Simperl}]{akhtar-etal-2024-chartcheck}
Mubashara Akhtar, Nikesh Subedi, Vivek Gupta, Sahar Tahmasebi, Oana Cocarascu, and Elena Simperl. 2024.
\newblock \href {https://doi.org/10.18653/v1/2024.findings-acl.828} {{C}hart{C}heck: Explainable fact-checking over real-world chart images}.
\newblock In \emph{Findings of the Association for Computational Linguistics: ACL 2024}, pages 13921--13937, Bangkok, Thailand. Association for Computational Linguistics.

\bibitem[{Augenstein et~al.(2019)Augenstein, Lioma, Wang, Chaves~Lima, Hansen, Hansen, and Simonsen}]{augenstein-etal-2019-multifc}
Isabelle Augenstein, Christina Lioma, Dongsheng Wang, Lucas Chaves~Lima, Casper Hansen, Christian Hansen, and Jakob~Grue Simonsen. 2019.
\newblock \href {https://doi.org/10.18653/v1/D19-1475} {{M}ulti{FC}: A real-world multi-domain dataset for evidence-based fact checking of claims}.
\newblock In \emph{Proceedings of the 2019 Conference on Empirical Methods in Natural Language Processing and the 9th International Joint Conference on Natural Language Processing (EMNLP-IJCNLP)}, pages 4685--4697, Hong Kong, China. Association for Computational Linguistics.

\bibitem[{Banerjee and Lavie(2005)}]{banerjee-lavie-2005-meteor}
Satanjeev Banerjee and Alon Lavie. 2005.
\newblock \href {https://aclanthology.org/W05-0909} {{METEOR}: An automatic metric for {MT} evaluation with improved correlation with human judgments}.
\newblock In \emph{Proceedings of the {ACL} Workshop on Intrinsic and Extrinsic Evaluation Measures for Machine Translation and/or Summarization}, pages 65--72, Ann Arbor, Michigan. Association for Computational Linguistics.

\bibitem[{Barr{\'{o}}n{-}Cede{\~{n}}o et~al.(2018)Barr{\'{o}}n{-}Cede{\~{n}}o, Elsayed, Suwaileh, M{\`{a}}rquez, Atanasova, Zaghouani, Kyuchukov, Martino, and Nakov}]{Barron-CedenoES18}
Alberto Barr{\'{o}}n{-}Cede{\~{n}}o, Tamer Elsayed, Reem Suwaileh, Llu{\'{\i}}s M{\`{a}}rquez, Pepa Atanasova, Wajdi Zaghouani, Spas Kyuchukov, Giovanni Da~San Martino, and Preslav Nakov. 2018.
\newblock Overview of the {CLEF-2018} checkthat! lab on automatic identification and verification of political claims. task 2: Factuality.
\newblock In \emph{Working Notes of {CLEF} 2018 - Conference and Labs of the Evaluation Forum, Avignon, France, September 10-14, 2018}, volume 2125 of \emph{{CEUR} Workshop Proceedings}. CEUR-WS.org.

\bibitem[{Celikyilmaz et~al.(2020)Celikyilmaz, Clark, and Gao}]{Celikyilmaz20}
Asli Celikyilmaz, Elizabeth Clark, and Jianfeng Gao. 2020.
\newblock \href {http://arxiv.org/abs/2006.14799} {Evaluation of text generation: {A} survey}.
\newblock \emph{CoRR}, abs/2006.14799.

\bibitem[{Chen et~al.(2024)Chen, Kim, Sriram, Durrett, and Choi}]{chen-etal-2024-complex}
Jifan Chen, Grace Kim, Aniruddh Sriram, Greg Durrett, and Eunsol Choi. 2024.
\newblock \href {https://doi.org/10.18653/v1/2024.naacl-long.196} {Complex claim verification with evidence retrieved in the wild}.
\newblock In \emph{Proceedings of the 2024 Conference of the North American Chapter of the Association for Computational Linguistics: Human Language Technologies (Volume 1: Long Papers)}, pages 3569--3587, Mexico City, Mexico. Association for Computational Linguistics.

\bibitem[{Devlin et~al.(2019)Devlin, Chang, Lee, and Toutanova}]{devlin-etal-2019-bert}
Jacob Devlin, Ming-Wei Chang, Kenton Lee, and Kristina Toutanova. 2019.
\newblock \href {https://doi.org/10.18653/v1/N19-1423} {{BERT}: Pre-training of deep bidirectional transformers for language understanding}.
\newblock In \emph{Proceedings of the 2019 Conference of the North {A}merican Chapter of the Association for Computational Linguistics: Human Language Technologies, Volume 1 (Long and Short Papers)}, pages 4171--4186, Minneapolis, Minnesota. Association for Computational Linguistics.

\bibitem[{Dubey et~al.(2024)Dubey, Jauhri, Pandey, Kadian, Al{-}Dahle, Letman, Mathur, Schelten, Yang, Fan, Goyal, Hartshorn, Yang, Mitra, Sravankumar, Korenev, Hinsvark, Rao, Zhang, Rodriguez, Gregerson, Spataru, Rozi{\`{e}}re, Biron, Tang, Chern, Caucheteux, Nayak, Bi, Marra, McConnell, Keller, Touret, Wu, Wong, Ferrer, Nikolaidis, Allonsius, Song, Pintz, Livshits, Esiobu, Choudhary, Mahajan, Garcia{-}Olano, Perino, Hupkes, Lakomkin, AlBadawy, Lobanova, Dinan, Smith, Radenovic, Zhang, Synnaeve, Lee, Anderson, Nail, Mialon, Pang, Cucurell, Nguyen, Korevaar, Xu, Touvron, Zarov, Ibarra, Kloumann, Misra, Evtimov, Copet, Lee, Geffert, Vranes, Park, Mahadeokar, Shah, van~der Linde, Billock, Hong, Lee, Fu, Chi, Huang, Liu, Wang, Yu, Bitton, Spisak, Park, Rocca, Johnstun, Saxe, Jia, Alwala, Upasani, Plawiak, Li, Heafield, Stone, and et~al.}]{Dubey-etal-2024}
Abhimanyu Dubey, Abhinav Jauhri, Abhinav Pandey, Abhishek Kadian, Ahmad Al{-}Dahle, Aiesha Letman, Akhil Mathur, Alan Schelten, Amy Yang, Angela Fan, Anirudh Goyal, Anthony Hartshorn, Aobo Yang, Archi Mitra, Archie Sravankumar, Artem Korenev, Arthur Hinsvark, Arun Rao, Aston Zhang, Aur{\'{e}}lien Rodriguez, Austen Gregerson, Ava Spataru, Baptiste Rozi{\`{e}}re, Bethany Biron, Binh Tang, Bobbie Chern, Charlotte Caucheteux, Chaya Nayak, Chloe Bi, Chris Marra, Chris McConnell, Christian Keller, Christophe Touret, Chunyang Wu, Corinne Wong, Cristian~Canton Ferrer, Cyrus Nikolaidis, Damien Allonsius, Daniel Song, Danielle Pintz, Danny Livshits, David Esiobu, Dhruv Choudhary, Dhruv Mahajan, Diego Garcia{-}Olano, Diego Perino, Dieuwke Hupkes, Egor Lakomkin, Ehab AlBadawy, Elina Lobanova, Emily Dinan, Eric~Michael Smith, Filip Radenovic, Frank Zhang, Gabriel Synnaeve, Gabrielle Lee, Georgia~Lewis Anderson, Graeme Nail, Gr{\'{e}}goire Mialon, Guan Pang, Guillem Cucurell, Hailey Nguyen, Hannah Korevaar, Hu~Xu, Hugo
  Touvron, Iliyan Zarov, Imanol~Arrieta Ibarra, Isabel~M. Kloumann, Ishan Misra, Ivan Evtimov, Jade Copet, Jaewon Lee, Jan Geffert, Jana Vranes, Jason Park, Jay Mahadeokar, Jeet Shah, Jelmer van~der Linde, Jennifer Billock, Jenny Hong, Jenya Lee, Jeremy Fu, Jianfeng Chi, Jianyu Huang, Jiawen Liu, Jie Wang, Jiecao Yu, Joanna Bitton, Joe Spisak, Jongsoo Park, Joseph Rocca, Joshua Johnstun, Joshua Saxe, Junteng Jia, Kalyan~Vasuden Alwala, Kartikeya Upasani, Kate Plawiak, Ke~Li, Kenneth Heafield, Kevin Stone, and et~al. 2024.
\newblock \href {https://doi.org/10.48550/ARXIV.2407.21783} {The llama 3 herd of models}.
\newblock \emph{CoRR}, abs/2407.21783.

\bibitem[{Durmus et~al.(2020)Durmus, He, and Diab}]{durmus-etal-2020-feqa}
Esin Durmus, He~He, and Mona Diab. 2020.
\newblock \href {https://doi.org/10.18653/v1/2020.acl-main.454} {{FEQA}: A question answering evaluation framework for faithfulness assessment in abstractive summarization}.
\newblock In \emph{Proceedings of the 58th Annual Meeting of the Association for Computational Linguistics}, pages 5055--5070, Online. Association for Computational Linguistics.

\bibitem[{Eyal et~al.(2019)Eyal, Baumel, and Elhadad}]{eyal-etal-2019-question}
Matan Eyal, Tal Baumel, and Michael Elhadad. 2019.
\newblock \href {https://doi.org/10.18653/v1/N19-1395} {Question answering as an automatic evaluation metric for news article summarization}.
\newblock In \emph{Proceedings of the 2019 Conference of the North {A}merican Chapter of the Association for Computational Linguistics: Human Language Technologies, Volume 1 (Long and Short Papers)}, pages 3938--3948, Minneapolis, Minnesota. Association for Computational Linguistics.

\bibitem[{Fleiss(1971)}]{fleiss1971measuring}
Joseph~L. Fleiss. 1971.
\newblock \href {https://doi.org/10.1037/h0031619} {Measuring nominal scale agreement among several raters}.
\newblock \emph{Psychological Bulletin}, 76(5):378--382.

\bibitem[{Fu et~al.(2023)Fu, Ng, Jiang, and Liu}]{FuNJL23}
Jinlan Fu, See{-}Kiong Ng, Zhengbao Jiang, and Pengfei Liu. 2023.
\newblock \href {https://doi.org/10.48550/ARXIV.2302.04166} {Gptscore: Evaluate as you desire}.
\newblock \emph{CoRR}, abs/2302.04166.

\bibitem[{Goodrich et~al.(2019)Goodrich, Rao, Liu, and Saleh}]{GoodrichRLS19}
Ben Goodrich, Vinay Rao, Peter~J. Liu, and Mohammad Saleh. 2019.
\newblock \href {https://doi.org/10.1145/3292500.3330955} {Assessing the factual accuracy of generated text}.
\newblock In \emph{Proceedings of the 25th {ACM} {SIGKDD} International Conference on Knowledge Discovery {\&} Data Mining, {KDD} 2019, Anchorage, AK, USA, August 4-8, 2019}, pages 166--175. {ACM}.

\bibitem[{Graves(2018)}]{graves2018a}
L~Graves. 2018.
\newblock Understanding the promise and limits of automated fact-checking.
\newblock Technical report.

\bibitem[{Graves(2017)}]{graves2017}
Lucas Graves. 2017.
\newblock \href {https://doi.org/10.1111/cccr.12163} {{Anatomy of a Fact Check: Objective Practice and the Contested Epistemology of Fact Checking}}.
\newblock \emph{Communication, Culture and Critique}, 10(3):518--537.

\bibitem[{Guo et~al.(2022)Guo, Schlichtkrull, and Vlachos}]{guo-etal-2022-survey}
Zhijiang Guo, Michael Schlichtkrull, and Andreas Vlachos. 2022.
\newblock \href {https://doi.org/10.1162/tacl_a_00454} {A survey on automated fact-checking}.
\newblock \emph{Transactions of the Association for Computational Linguistics}, 10:178--206.

\bibitem[{He et~al.(2021)He, Liu, Gao, and Chen}]{HeLGC21}
Pengcheng He, Xiaodong Liu, Jianfeng Gao, and Weizhu Chen. 2021.
\newblock \href {https://openreview.net/forum?id=XPZIaotutsD} {Deberta: decoding-enhanced bert with disentangled attention}.
\newblock In \emph{9th International Conference on Learning Representations, {ICLR} 2021, Virtual Event, Austria, May 3-7, 2021}. OpenReview.net.

\bibitem[{Horne et~al.(2018)Horne, Khedr, and Adali}]{HorneKA18}
Benjamin~D. Horne, Sara Khedr, and Sibel Adali. 2018.
\newblock \href {https://aaai.org/ocs/index.php/ICWSM/ICWSM18/paper/view/17796} {Sampling the news producers: {A} large news and feature data set for the study of the complex media landscape}.
\newblock In \emph{Proceedings of the Twelfth International Conference on Web and Social Media, {ICWSM} 2018, Stanford, California, USA, June 25-28, 2018}, pages 518--527. {AAAI} Press.

\bibitem[{Huang and Zhang(2021)}]{huang-zhang-2021-evaluation}
Nannan Huang and Xiuzhen Zhang. 2021.
\newblock \href {https://aclanthology.org/2021.alta-1.9} {Evaluation of review summaries via question-answering}.
\newblock In \emph{Proceedings of the The 19th Annual Workshop of the Australasian Language Technology Association}, pages 87--96, Online. Australasian Language Technology Association.

\bibitem[{Ji et~al.(2023)Ji, Lee, Frieske, Yu, Su, Xu, Ishii, Bang, Madotto, and Fung}]{JiLFYSXIBMF23}
Ziwei Ji, Nayeon Lee, Rita Frieske, Tiezheng Yu, Dan Su, Yan Xu, Etsuko Ishii, Yejin Bang, Andrea Madotto, and Pascale Fung. 2023.
\newblock \href {https://doi.org/10.1145/3571730} {Survey of hallucination in natural language generation}.
\newblock \emph{{ACM} Comput. Surv.}, 55(12):248:1--248:38.

\bibitem[{Jiang et~al.(2020)Jiang, Bordia, Zhong, Dognin, Singh, and Bansal}]{jiang-etal-2020-hover}
Yichen Jiang, Shikha Bordia, Zheng Zhong, Charles Dognin, Maneesh Singh, and Mohit Bansal. 2020.
\newblock \href {https://doi.org/10.18653/v1/2020.findings-emnlp.309} {{H}o{V}er: A dataset for many-hop fact extraction and claim verification}.
\newblock In \emph{Findings of the Association for Computational Linguistics: EMNLP 2020}, pages 3441--3460, Online. Association for Computational Linguistics.

\bibitem[{Kane et~al.(2020)Kane, Kocyigit, Abdalla, Ajanoh, and Coulibali}]{kane-etal-2020-nubia}
Hassan Kane, Muhammed~Yusuf Kocyigit, Ali Abdalla, Pelkins Ajanoh, and Mohamed Coulibali. 2020.
\newblock \href {https://aclanthology.org/2020.evalnlgeval-1.4} {{NUBIA}: {N}e{U}ral based interchangeability assessor for text generation}.
\newblock In \emph{Proceedings of the 1st Workshop on Evaluating NLG Evaluation}, pages 28--37, Online (Dublin, Ireland). Association for Computational Linguistics.

\bibitem[{Krippendorff(2004)}]{krippendorff2004content}
Klaus Krippendorff. 2004.
\newblock Content analysis: An introduction to its methodology.

\bibitem[{Lin(2004)}]{lin-2004-rouge}
Chin-Yew Lin. 2004.
\newblock \href {https://aclanthology.org/W04-1013} {{ROUGE}: A package for automatic evaluation of summaries}.
\newblock In \emph{Text Summarization Branches Out}, pages 74--81, Barcelona, Spain. Association for Computational Linguistics.

\bibitem[{Liu et~al.(2022)Liu, Swayamdipta, Smith, and Choi}]{liu-etal-2022-wanli}
Alisa Liu, Swabha Swayamdipta, Noah~A. Smith, and Yejin Choi. 2022.
\newblock \href {https://doi.org/10.18653/v1/2022.findings-emnlp.508} {{WANLI}: Worker and {AI} collaboration for natural language inference dataset creation}.
\newblock In \emph{Findings of the Association for Computational Linguistics: EMNLP 2022}, pages 6826--6847, Abu Dhabi, United Arab Emirates. Association for Computational Linguistics.

\bibitem[{McCoy et~al.(2019)McCoy, Pavlick, and Linzen}]{mccoy-etal-2019-right}
Tom McCoy, Ellie Pavlick, and Tal Linzen. 2019.
\newblock \href {https://doi.org/10.18653/v1/P19-1334} {Right for the wrong reasons: Diagnosing syntactic heuristics in natural language inference}.
\newblock In \emph{Proceedings of the 57th Annual Meeting of the Association for Computational Linguistics}, pages 3428--3448, Florence, Italy. Association for Computational Linguistics.

\bibitem[{Min et~al.(2023)Min, Krishna, Lyu, Lewis, Yih, Koh, Iyyer, Zettlemoyer, and Hajishirzi}]{min-etal-2023-factscore}
Sewon Min, Kalpesh Krishna, Xinxi Lyu, Mike Lewis, Wen-tau Yih, Pang Koh, Mohit Iyyer, Luke Zettlemoyer, and Hannaneh Hajishirzi. 2023.
\newblock \href {https://doi.org/10.18653/v1/2023.emnlp-main.741} {{FA}ct{S}core: Fine-grained atomic evaluation of factual precision in long form text generation}.
\newblock In \emph{Proceedings of the 2023 Conference on Empirical Methods in Natural Language Processing}, pages 12076--12100, Singapore. Association for Computational Linguistics.

\bibitem[{Nema and Khapra(2018)}]{nema-khapra-2018-towards}
Preksha Nema and Mitesh~M. Khapra. 2018.
\newblock \href {https://doi.org/10.18653/v1/D18-1429} {Towards a better metric for evaluating question generation systems}.
\newblock In \emph{Proceedings of the 2018 Conference on Empirical Methods in Natural Language Processing}, pages 3950--3959, Brussels, Belgium. Association for Computational Linguistics.

\bibitem[{Nie et~al.(2019)Nie, Chen, and Bansal}]{nie2019combining}
Yixin Nie, Haonan Chen, and Mohit Bansal. 2019.
\newblock Combining fact extraction and verification with neural semantic matching networks.
\newblock In \emph{Association for the Advancement of Artificial Intelligence ({AAAI})}.

\bibitem[{Nie et~al.(2020)Nie, Williams, Dinan, Bansal, Weston, and Kiela}]{nie-etal-2020-adversarial}
Yixin Nie, Adina Williams, Emily Dinan, Mohit Bansal, Jason Weston, and Douwe Kiela. 2020.
\newblock \href {https://doi.org/10.18653/v1/2020.acl-main.441} {Adversarial {NLI}: A new benchmark for natural language understanding}.
\newblock In \emph{Proceedings of the 58th Annual Meeting of the Association for Computational Linguistics}, pages 4885--4901, Online. Association for Computational Linguistics.

\bibitem[{N{\o}rregaard and Derczynski(2021)}]{norregaard-derczynski-2021-danfever}
Jeppe N{\o}rregaard and Leon Derczynski. 2021.
\newblock \href {https://aclanthology.org/2021.nodalida-main.47} {{D}an{FEVER}: claim verification dataset for {D}anish}.
\newblock In \emph{Proceedings of the 23rd Nordic Conference on Computational Linguistics (NoDaLiDa)}, pages 422--428, Reykjavik, Iceland (Online). Link{\"o}ping University Electronic Press, Sweden.

\bibitem[{OpenAI(2023)}]{GPT4}
OpenAI. 2023.
\newblock \href {https://doi.org/10.48550/ARXIV.2303.08774} {{GPT-4} technical report}.
\newblock \emph{CoRR}, abs/2303.08774.

\bibitem[{OpenAI(2024)}]{GPT4o-introduction}
GPT4o OpenAI. 2024.
\newblock Blog post: Hello gpt-4o.
\newblock \url{https://openai.com/index/hello-gpt-4o/}.
\newblock Accessed: 2024-10-23.

\bibitem[{Papineni et~al.(2002)Papineni, Roukos, Ward, and Zhu}]{papineni-etal-2002-bleu}
Kishore Papineni, Salim Roukos, Todd Ward, and Wei-Jing Zhu. 2002.
\newblock \href {https://doi.org/10.3115/1073083.1073135} {{B}leu: a method for automatic evaluation of machine translation}.
\newblock In \emph{Proceedings of the 40th Annual Meeting of the Association for Computational Linguistics}, pages 311--318, Philadelphia, Pennsylvania, USA. Association for Computational Linguistics.

\bibitem[{Parrish et~al.(2021)Parrish, Huang, Agha, Lee, Nangia, Warstadt, Aggarwal, Allaway, Linzen, and Bowman}]{parrish-etal-2021-putting-linguist}
Alicia Parrish, William Huang, Omar Agha, Soo-Hwan Lee, Nikita Nangia, Alexia Warstadt, Karmanya Aggarwal, Emily Allaway, Tal Linzen, and Samuel~R. Bowman. 2021.
\newblock \href {https://doi.org/10.18653/v1/2021.findings-emnlp.421} {Does putting a linguist in the loop improve {NLU} data collection?}
\newblock In \emph{Findings of the Association for Computational Linguistics: EMNLP 2021}, pages 4886--4901, Punta Cana, Dominican Republic. Association for Computational Linguistics.

\bibitem[{Pearson(1896)}]{Pearson1896}
Karl Pearson. 1896.
\newblock Vii. mathematical contributions to the theory of evolution.—iii. regression, heredity, and panmixia.
\newblock \emph{Philosophical Transactions of the Royal Society of London. Series A, containing papers of a mathematical or physical character}, (187):253--318.

\bibitem[{Reid et~al.(2024)Reid, Savinov, Teplyashin, Lepikhin, Lillicrap, Alayrac, Soricut, Lazaridou, Firat, Schrittwieser, Antonoglou, Anil, Borgeaud, Dai, Millican, Dyer, Glaese, Sottiaux, Lee, Viola, Reynolds, Xu, Molloy, Chen, Isard, Barham, Hennigan, McIlroy, Johnson, Schalkwyk, Collins, Rutherford, Moreira, Ayoub, Goel, Meyer, Thornton, Yang, Michalewski, Abbas, Schucher, Anand, Ives, Keeling, Lenc, Haykal, Shakeri, Shyam, Chowdhery, Ring, Spencer, Sezener, and et~al.}]{Gemini1.5}
Machel Reid, Nikolay Savinov, Denis Teplyashin, Dmitry Lepikhin, Timothy~P. Lillicrap, Jean{-}Baptiste Alayrac, Radu Soricut, Angeliki Lazaridou, Orhan Firat, Julian Schrittwieser, Ioannis Antonoglou, Rohan Anil, Sebastian Borgeaud, Andrew~M. Dai, Katie Millican, Ethan Dyer, Mia Glaese, Thibault Sottiaux, Benjamin Lee, Fabio Viola, Malcolm Reynolds, Yuanzhong Xu, James Molloy, Jilin Chen, Michael Isard, Paul Barham, Tom Hennigan, Ross McIlroy, Melvin Johnson, Johan Schalkwyk, Eli Collins, Eliza Rutherford, Erica Moreira, Kareem Ayoub, Megha Goel, Clemens Meyer, Gregory Thornton, Zhen Yang, Henryk Michalewski, Zaheer Abbas, Nathan Schucher, Ankesh Anand, Richard Ives, James Keeling, Karel Lenc, Salem Haykal, Siamak Shakeri, Pranav Shyam, Aakanksha Chowdhery, Roman Ring, Stephen Spencer, Eren Sezener, and et~al. 2024.
\newblock \href {https://doi.org/10.48550/ARXIV.2403.05530} {Gemini 1.5: Unlocking multimodal understanding across millions of tokens of context}.
\newblock \emph{CoRR}, abs/2403.05530.

\bibitem[{Ribeiro et~al.(2020)Ribeiro, Wu, Guestrin, and Singh}]{ribeiro-etal-2020-beyond}
Marco~Tulio Ribeiro, Tongshuang Wu, Carlos Guestrin, and Sameer Singh. 2020.
\newblock \href {https://doi.org/10.18653/v1/2020.acl-main.442} {Beyond accuracy: Behavioral testing of {NLP} models with {C}heck{L}ist}.
\newblock In \emph{Proceedings of the 58th Annual Meeting of the Association for Computational Linguistics}, pages 4902--4912, Online. Association for Computational Linguistics.

\bibitem[{Sai et~al.(2023)Sai, Mohankumar, and Khapra}]{SaiMK23}
Ananya~B. Sai, Akash~Kumar Mohankumar, and Mitesh~M. Khapra. 2023.
\newblock \href {https://doi.org/10.1145/3485766} {A survey of evaluation metrics used for {NLG} systems}.
\newblock \emph{{ACM} Comput. Surv.}, 55(2):26:1--26:39.

\bibitem[{Sathe et~al.(2020)Sathe, Ather, Le, Perry, and Park}]{sathe-etal-2020-automated}
Aalok Sathe, Salar Ather, Tuan~Manh Le, Nathan Perry, and Joonsuk Park. 2020.
\newblock \href {https://aclanthology.org/2020.lrec-1.849} {Automated fact-checking of claims from {W}ikipedia}.
\newblock In \emph{Proceedings of the Twelfth Language Resources and Evaluation Conference}, pages 6874--6882, Marseille, France. European Language Resources Association.

\bibitem[{Schlichtkrull et~al.(2024{\natexlab{a}})Schlichtkrull, Chen, Whitehouse, Deng, Akhtar, Aly, Guo, Christodoulopoulos, Cocarascu, Mittal, Thorne, and Vlachos}]{schlichtkrull-etal-2024-automated}
Michael Schlichtkrull, Yulong Chen, Chenxi Whitehouse, Zhenyun Deng, Mubashara Akhtar, Rami Aly, Zhijiang Guo, Christos Christodoulopoulos, Oana Cocarascu, Arpit Mittal, James Thorne, and Andreas Vlachos. 2024{\natexlab{a}}.
\newblock \href {https://doi.org/10.18653/v1/2024.fever-1.1} {The automated verification of textual claims ({AV}eri{T}e{C}) shared task}.
\newblock In \emph{Proceedings of the Seventh Fact Extraction and VERification Workshop (FEVER)}, pages 1--26, Miami, Florida, USA. Association for Computational Linguistics.

\bibitem[{Schlichtkrull et~al.(2024{\natexlab{b}})Schlichtkrull, Chen, Whitehouse, Deng, Akhtar, Aly, Guo, Christodoulopoulos, Cocarascu, Mittal, Thorne, and Vlachos}]{schlichtkrull2024automatedverificationtextualclaims}
Michael Schlichtkrull, Yulong Chen, Chenxi Whitehouse, Zhenyun Deng, Mubashara Akhtar, Rami Aly, Zhijiang Guo, Christos Christodoulopoulos, Oana Cocarascu, Arpit Mittal, James Thorne, and Andreas Vlachos. 2024{\natexlab{b}}.
\newblock \href {http://arxiv.org/abs/2410.23850} {The automated verification of textual claims (averitec) shared task}.
\newblock \emph{CoRR}, abs/2410.23850.

\bibitem[{Schlichtkrull et~al.(2023)Schlichtkrull, Guo, and Vlachos}]{schlichtkrullGV2023}
Michael~Sejr Schlichtkrull, Zhijiang Guo, and Andreas Vlachos. 2023.
\newblock \href {https://doi.org/10.48550/arXiv.2305.13117} {Averitec: {A} dataset for real-world claim verification with evidence from the web}.
\newblock \emph{CoRR}, abs/2305.13117.

\bibitem[{Schuster et~al.(2021)Schuster, Fisch, and Barzilay}]{schuster-etal-2021-get}
Tal Schuster, Adam Fisch, and Regina Barzilay. 2021.
\newblock \href {https://doi.org/10.18653/v1/2021.naacl-main.52} {Get your vitamin {C}! robust fact verification with contrastive evidence}.
\newblock In \emph{Proceedings of the 2021 Conference of the North American Chapter of the Association for Computational Linguistics: Human Language Technologies}, pages 624--643, Online. Association for Computational Linguistics.

\bibitem[{Scialom et~al.(2019)Scialom, Lamprier, Piwowarski, and Staiano}]{scialom-etal-2019-answers}
Thomas Scialom, Sylvain Lamprier, Benjamin Piwowarski, and Jacopo Staiano. 2019.
\newblock \href {https://doi.org/10.18653/v1/D19-1320} {Answers unite! unsupervised metrics for reinforced summarization models}.
\newblock In \emph{Proceedings of the 2019 Conference on Empirical Methods in Natural Language Processing and the 9th International Joint Conference on Natural Language Processing (EMNLP-IJCNLP)}, pages 3246--3256, Hong Kong, China. Association for Computational Linguistics.

\bibitem[{Sellam et~al.(2020)Sellam, Das, and Parikh}]{sellam-etal-2020-bleurt}
Thibault Sellam, Dipanjan Das, and Ankur Parikh. 2020.
\newblock \href {https://doi.org/10.18653/v1/2020.acl-main.704} {{BLEURT}: Learning robust metrics for text generation}.
\newblock In \emph{Proceedings of the 58th Annual Meeting of the Association for Computational Linguistics}, pages 7881--7892, Online. Association for Computational Linguistics.

\bibitem[{Shahi and Nandini(2020)}]{ShahiN20}
Gautam~Kishore Shahi and Durgesh Nandini. 2020.
\newblock \href {https://doi.org/10.36190/2020.14} {Fakecovid- {A} multilingual cross-domain fact check news dataset for {COVID-19}}.
\newblock In \emph{Workshop Proceedings of the 14th International {AAAI} Conference on Web and Social Media, {ICWSM} 2020 Workshops, Atlanta, Georgia, {USA} [virtual], June 8, 2020}.

\bibitem[{Spearman(1987)}]{Spearman1987}
C.~Spearman. 1987.
\newblock \href {http://www.jstor.org/stable/1422689} {The proof and measurement of association between two things}.
\newblock \emph{The American Journal of Psychology}, 100(3/4):441--471.

\bibitem[{Thorne et~al.(2018{\natexlab{a}})Thorne, Vlachos, Christodoulopoulos, and Mittal}]{thorne-etal-2018-fever}
James Thorne, Andreas Vlachos, Christos Christodoulopoulos, and Arpit Mittal. 2018{\natexlab{a}}.
\newblock \href {https://doi.org/10.18653/v1/N18-1074} {{FEVER}: a large-scale dataset for fact extraction and {VER}ification}.
\newblock In \emph{Proceedings of the 2018 Conference of the North {A}merican Chapter of the Association for Computational Linguistics: Human Language Technologies, Volume 1 (Long Papers)}, pages 809--819, New Orleans, Louisiana. Association for Computational Linguistics.

\bibitem[{Thorne et~al.(2018{\natexlab{b}})Thorne, Vlachos, Cocarascu, Christodoulopoulos, and Mittal}]{thorne-etal-2018-fact}
James Thorne, Andreas Vlachos, Oana Cocarascu, Christos Christodoulopoulos, and Arpit Mittal. 2018{\natexlab{b}}.
\newblock \href {https://doi.org/10.18653/v1/W18-5501} {The fact extraction and {VER}ification ({FEVER}) shared task}.
\newblock In \emph{Proceedings of the First Workshop on Fact Extraction and {VER}ification ({FEVER})}, pages 1--9, Brussels, Belgium. Association for Computational Linguistics.

\bibitem[{Wei et~al.(2022)Wei, Wang, Schuurmans, Bosma, Ichter, Xia, Chi, Le, and Zhou}]{Wei0SBIXCLZ22}
Jason Wei, Xuezhi Wang, Dale Schuurmans, Maarten Bosma, Brian Ichter, Fei Xia, Ed~H. Chi, Quoc~V. Le, and Denny Zhou. 2022.
\newblock Chain-of-thought prompting elicits reasoning in large language models.
\newblock In \emph{Advances in Neural Information Processing Systems 35: Annual Conference on Neural Information Processing Systems 2022, NeurIPS 2022, New Orleans, LA, USA, November 28 - December 9, 2022}.

\bibitem[{Williams et~al.(2018)Williams, Nangia, and Bowman}]{williams-etal-2018-broad}
Adina Williams, Nikita Nangia, and Samuel Bowman. 2018.
\newblock \href {https://doi.org/10.18653/v1/N18-1101} {A broad-coverage challenge corpus for sentence understanding through inference}.
\newblock In \emph{Proceedings of the 2018 Conference of the North {A}merican Chapter of the Association for Computational Linguistics: Human Language Technologies, Volume 1 (Long Papers)}, pages 1112--1122, New Orleans, Louisiana. Association for Computational Linguistics.

\bibitem[{Zhang et~al.(2020)Zhang, Kishore, Wu, Weinberger, and Artzi}]{ZhangKWWA20}
Tianyi Zhang, Varsha Kishore, Felix Wu, Kilian~Q. Weinberger, and Yoav Artzi. 2020.
\newblock \href {https://openreview.net/forum?id=SkeHuCVFDr} {Bertscore: Evaluating text generation with {BERT}}.
\newblock In \emph{8th International Conference on Learning Representations, {ICLR} 2020, Addis Ababa, Ethiopia, April 26-30, 2020}. OpenReview.net.

\bibitem[{Zhao et~al.(2019)Zhao, Peyrard, Liu, Gao, Meyer, and Eger}]{zhao-etal-2019-moverscore}
Wei Zhao, Maxime Peyrard, Fei Liu, Yang Gao, Christian~M. Meyer, and Steffen Eger. 2019.
\newblock \href {https://doi.org/10.18653/v1/D19-1053} {{M}over{S}core: Text generation evaluating with contextualized embeddings and earth mover distance}.
\newblock In \emph{Proceedings of the 2019 Conference on Empirical Methods in Natural Language Processing and the 9th International Joint Conference on Natural Language Processing (EMNLP-IJCNLP)}, pages 563--578, Hong Kong, China. Association for Computational Linguistics.

\end{thebibliography}
\bibliographystyle{acl_natbib}

% \iftaclpubformat

\onecolumn

\appendix

\section{Evaluation with adversarial perturbations}

\begin{table*}[ht]
    \centering
    \scalebox{0.88}{
    \begin{tabular}{p{0.02\linewidth} p{0.16\linewidth} p{0.75\linewidth}} \toprule
        \# & Test Type & Description \\\midrule
        1 & Completeness & Drops substantial parts of the evidence, causing the veracity label to change, i.e., $y \in \{\text{support}; \text{refute}\}$ to \textit{not enough information (NEI)}. We expect Ev\textsuperscript{2}R scorers to react to these changes with a reduction in scores. \\
        2 & Random Shuffle & Randomly shuffles all words occurring in an evidence such that the resulting evidence text is not understandable. \\
        3 & Fluency & Semantics-preserving tests that $(a)$ introduce typos or $(b)$ drop stop words. \\
        4 & Invariance & Introducing invariant changes to the evidence data: $(a)$ changing numbers to text (e.g., $2$ to \textit{two}); $(b)$ changing numbers as text to numerals (e.g., \textit{two} to $2$); $(c)$ replacing words in evidence by their synonyms; $(d)$ introducing contractions (e.g., \textit{we're} instead of \textit{we are}). \\
        5 & Noise & Adds noise by inserting a random sentence from another evidence set in the dataset. \\
        6 & Redundancy & Introduces redundancy by duplicating sentences or words within sentences. \\
        7 & Argument Structure & Changes the order of evidence sentences to disrupt the logical flow of arguments. \\
    \bottomrule
    \end{tabular}}
    \caption{Categories of adversarial tests with descriptions.}
    \label{table:adversarial_test_dimensions}
\end{table*}

\section{Additional Evaluation Dimensions}

As an additional data source for the evaluation of scorers, we collected ratings for 100 randomly selected test data instances from the AVeriTeC test set~\citep{schlichtkrullGV2023}, together with the evidence retrieved through the AVeriTeC baseline system as described in \citet{schlichtkrullGV2023}. Each test sample was evaluated by two annotators, who were either computer science graduate students or postdoctoral researchers familiar with AFC research.
This resulted in 200 annotations while the overall rating for each annotated sample and dimension was calculated as the average of both annotations. If the annotators disagreed on the verdict assigned to the retrieved evidence, hence interpreted the relationship between retrieved evidence and claims differently, we assigned the data instance to a third annotator and used majority voting to determine the final verdict.

\begin{table*}[!htbp]
\centering
\scalebox{0.85}{
\begin{tabular}{l | c c | c c | c c | c c | c c }
\hline
\textbf{Scorer} & \multicolumn{2}{c|}{\bf Coverage} & \multicolumn{2}{c|}{\bf Coherence} & \multicolumn{2}{c|}{\bf Repetition} & \multicolumn{2}{c|}{\bf Consistency} & \multicolumn{2}{c}{\bf Relevance} \\
 & $\rho$ & $r$ & $\rho$ & $r$ & $\rho$ & $r$ & $\rho$ & $r$ & $\rho$ & $r$ \\
\hline
\multicolumn{11}{l}{ Prompt Scorer GPT$4o$} \\
\hline
Ref-less & .237 & .261 & .201 & .254 & -.006 & -.009 & .112 & .118 & .292 & .360 \\
\textcolor{darkgray}{Ref-based (prec)} & \textcolor{darkgray}{.277} & \textcolor{darkgray}{.270} & \textcolor{darkgray}{.185} & \textcolor{darkgray}{.191} & \textcolor{darkgray}{-.089} & \textcolor{darkgray}{-.090} & \textcolor{darkgray}{.218} & \textcolor{darkgray}{.242} & \textcolor{darkgray}{.316} & \textcolor{darkgray}{.313} \\
\textcolor{darkgray}{Ref-based (recall)} & \textcolor{darkgray}{.339} & \textcolor{darkgray}{.334} & \textcolor{darkgray}{.199} & \textcolor{darkgray}{.197} & \textcolor{darkgray}{-.140} & \textcolor{darkgray}{-.105} & \textcolor{darkgray}{.187} & \textcolor{darkgray}{.190} & \textcolor{darkgray}{.277} & \textcolor{darkgray}{.269} \\
Ref-based ($F_1$) & .306 & .298 & .192 & .194 & -.109 & -.097 & .201 & .213 & .295 & .289 \\
Proxy-ref & .253 & .306 & .142 & .201 & .007 & -.011 & .133 & .140 & .182 & .221 \\
\hline
\multicolumn{11}{l}{ Prompt Scorer Gemini-Pro} \\
\hline
Ref-less & .287 & .296 & .208 & .246 & .012 & .010 & .092 & .110 & .203 & .263 \\
\textcolor{darkgray}{Ref-based (prec)} & \textcolor{darkgray}{.304} & \textcolor{darkgray}{.254} & \textcolor{darkgray}{.186} & \textcolor{darkgray}{.163} & \textcolor{darkgray}{-.094} & \textcolor{darkgray}{-.115} & \textcolor{darkgray}{.193} & \textcolor{darkgray}{.197} & \textcolor{darkgray}{.270} & \textcolor{darkgray}{.264} \\
\textcolor{darkgray}{Ref-based (recall)} & \textcolor{darkgray}{.398} & \textcolor{darkgray}{.380} & \textcolor{darkgray}{.282} & \textcolor{darkgray}{.268} & \textcolor{darkgray}{-.168} & \textcolor{darkgray}{-.164} & \textcolor{darkgray}{.199} & \textcolor{darkgray}{.205} & \textcolor{darkgray}{.248} & \textcolor{darkgray}{.248} \\
Ref-based (F$_1$) & \textcolor{skyblue}{.345} & .304 & .224 & .203 & -.120 & -.135 & .196 & .201 & .259 & .256 \\
\hline
\multicolumn{11}{l}{ Prompt Scorer Gemini-Flash} \\
\hline
Ref-less & .275 & .287 & \textcolor{skyblue}{.256} & \textcolor{orangebrown}{.269} & .027 & .019 & .147 & .146 & .287 & .333 \\
\textcolor{darkgray}{Ref-based (prec)} & \textcolor{darkgray}{.086} & \textcolor{darkgray}{.232} & \textcolor{darkgray}{.105} & \textcolor{darkgray}{-.004} & \textcolor{darkgray}{-.481} & \textcolor{darkgray}{ -.450} & \textcolor{darkgray}{.337} & \textcolor{darkgray}{.181} & \textcolor{darkgray}{.490} & \textcolor{darkgray}{.368} \\
\textcolor{darkgray}{Ref-based (recall)} & \textcolor{darkgray}{.132} & \textcolor{darkgray}{.270} & \textcolor{darkgray}{.103} & \textcolor{darkgray}{.106} & \textcolor{darkgray}{-.577} & \textcolor{darkgray}{-.619} & \textcolor{darkgray}{.330} & \textcolor{darkgray}{.291} & \textcolor{darkgray}{.527} & \textcolor{darkgray}{.531} \\
Ref-based (F$_1$) & .104 & .249 & .104 & -.008 & \textcolor{skyblue}{-.524} & \textcolor{skyblue}{-.521} & \textcolor{skyblue}{.333} & .223 & \textcolor{skyblue}{.508} & \textcolor{skyblue}{.435} \\
\hline
\multicolumn{11}{l}{ Prompt Scorer Llama 3.1} \\
\hline
Ref-less & .297 & .286 & .222 & .210 & -.011 & -.008 & .134 & .116 & .227 & .250 \\
\textcolor{darkgray}{Ref-based (prec)} & \textcolor{darkgray}{.341} & \textcolor{darkgray}{.336} & \textcolor{darkgray}{.178} & \textcolor{darkgray}{.181} & \textcolor{darkgray}{-.120} & \textcolor{darkgray}{-.122} & \textcolor{darkgray}{.080} & \textcolor{darkgray}{.111} & \textcolor{darkgray}{.131} & \textcolor{darkgray}{.172} \\
\textcolor{darkgray}{Ref-based (recall)} & \textcolor{darkgray}{.299} & \textcolor{darkgray}{.313} & \textcolor{darkgray}{.174} & \textcolor{darkgray}{.179} & \textcolor{darkgray}{-.060} & \textcolor{darkgray}{-.048} & \textcolor{darkgray}{.148} & \textcolor{darkgray}{.143} & \textcolor{darkgray}{.175} & \textcolor{darkgray}{.187} \\
Ref-based (F$_1$) & .318 & .324 & .176 & .180 & -.080 & -.069 & .104 & .125 & .150 & .179 \\
\hline
\multicolumn{11}{l}{ Trained Scorer} \\
\hline
Ref-based & .116 & .033 & .091 & .063 & -.133 & .024 & .057 & .059 & .145 & .110 \\
Proxy-ref & .338 & \textcolor{skyblue}{.348} & \textcolor{orangebrown}{.230} & \textcolor{skyblue}{.286} & -.011 & -.057 & .293 & \textcolor{skyblue}{.329} & .298 & .374 \\
\hline
\multicolumn{11}{l}{ Baselines} \\
\hline
RougeL & .150 & .169 & .180 & .190 & .057 & .040 & .124 & .131 & .086 & .099 \\
BLEU & .236 & .184 & .180 & .166 & -.144 & -.039 & .040 & .038 & .107 & .079 \\
Meteor & .229 & .240 & .192 & .191 & -.150 & -.132 & .061 & .064 & .062 & .076 \\
H-METEOR & .005 & -.024 & .076 & .057 & .117 & .025 & .039 & .024 & .008 & .003 \\
\hline
\hline
\multicolumn{11}{l}{\textbf{Weighted Scorer ($\alpha = 0.5$)}} \\[0.2em]
\hline
GPT-4o / Proxy-ref   & .321 & .323 & .211 & .240 & -.060 & -.077 & .247 & .271 & .297 & .332 \\
Gemini-Pro / Proxy-ref & \textcolor{orangebrown}{.341} & .326 & .227 & .244 & -.066 & -.096 & .245 & .265 & .278 & .315 \\
Gemini-Flash / Proxy-ref & .221 & .299 & .167 & .147 & \textcolor{orangebrown}{-.267} & \textcolor{orangebrown}{-.289} & \textcolor{orangebrown}{.313} & \textcolor{orangebrown}{.276} & \textcolor{orangebrown}{.403} & \textcolor{orangebrown}{.404} \\
Llama 3.1 / Proxy-ref  & .328 & \textcolor{orangebrown}{.336} & .203 & .233 & -.046 & -.063 & .199 & .227 & .224 & .277 \\
\hline
\hline
\end{tabular}}
\caption{\label{tab:results_averitec_shared_task} Correlation between human-rated \textbf{AVeriTeC shared task submissions} and proposed scorers, across five quality dimensions. \textcolor{skyblue}{Highest scores per column} are colored blue and \textcolor{orangebrown}{second-highest values} highlighted with brown color. \textcolor{darkgray}{Reference-based precision and recall} scores which are calculated to compute the final reference-based $F_1$ score, are colored in gray.}
\vspace{-1em}
\end{table*}

Table \Cref{table:results_averitec_shared_task_trimmed_full_dimensions} given an overview over correlation between scorers and human ratings on three additional dimensions, shortly outlined below. 

\noindent \textbf{$(1)$ Coherence} captures whether the retrieved evidence is coherent, i.e., all sentences are connected sensibly and the evidence makes sense as a whole. An important aspect of fact-checking is presenting the verification process to the reader in a way that is comprehensible. Hence, well-structured and coherent evidence supports understanding of the verification process and the rationale behind the resulting verdict.

\noindent \textbf{$(2)$ Repetition} evaluates whether the retrieved evidence exhibits repetition of its content. As mentioned earlier, supporting readers’ understanding of the fact-checking process can be enhanced through avoiding repetitive content in the evidence. This can further maintain the reader's focus on unique information in the evidence text, leading to better understanding of the resulting conclusion, i.e., the verdict.

\noindent \textbf{$(3)$ Consistency} assesses whether the retrieved evidence is logically consistent in itself, since consistency in evidence improves the reliability of the fact-checking process and trust among readers.

\begin{table*}[!htbp]
\centering
\scalebox{0.85}{
\begin{tabular}{l | c c | c c | c c | c c | c c }
\hline
\textbf{Scorer} & \multicolumn{2}{c|}{\bf Coverage} & \multicolumn{2}{c|}{\bf Coherence} & \multicolumn{2}{c|}{\bf Repetition} & \multicolumn{2}{c|}{\bf Consistency} & \multicolumn{2}{c}{\bf Relevance} \\
 & $\rho$ & $r$ & $\rho$ & $r$ & $\rho$ & $r$ & $\rho$ & $r$ & $\rho$ & $r$ \\
\hline
\multicolumn{11}{l}{ Prompt Scorer GPT$4o$} \\
\hline
Ref-less & .237 & .261 & .201 & .254 & -.006 & -.009 & .112 & .118 & .292 & .360 \\
\textcolor{darkgray}{Ref-based (prec)} & \textcolor{darkgray}{.277} & \textcolor{darkgray}{.270} & \textcolor{darkgray}{.185} & \textcolor{darkgray}{.191} & \textcolor{darkgray}{-.089} & \textcolor{darkgray}{-.090} & \textcolor{darkgray}{.218} & \textcolor{darkgray}{.242} & \textcolor{darkgray}{.316} & \textcolor{darkgray}{.313} \\
\textcolor{darkgray}{Ref-based (recall)} & \textcolor{darkgray}{.339} & \textcolor{darkgray}{.334} & \textcolor{darkgray}{.199} & \textcolor{darkgray}{.197} & \textcolor{darkgray}{-.140} & \textcolor{darkgray}{-.105} & \textcolor{darkgray}{.187} & \textcolor{darkgray}{.190} & \textcolor{darkgray}{.277} & \textcolor{darkgray}{.269} \\
Ref-based ($F_1$) & .306 & .298 & .192 & .194 & -.109 & -.097 & .201 & .213 & .295 & .289 \\
Proxy-ref & .253 & .306 & .142 & .201 & .007 & -.011 & .133 & .140 & .182 & .221 \\
\hline
\multicolumn{11}{l}{ Prompt Scorer Gemini-Pro} \\
\hline
Ref-less & .287 & .296 & .208 & .246 & .012 & .010 & .092 & .110 & .203 & .263 \\
\textcolor{darkgray}{Ref-based (prec)} & \textcolor{darkgray}{.304} & \textcolor{darkgray}{.254} & \textcolor{darkgray}{.186} & \textcolor{darkgray}{.163} & \textcolor{darkgray}{-.094} & \textcolor{darkgray}{-.115} & \textcolor{darkgray}{.193} & \textcolor{darkgray}{.197} & \textcolor{darkgray}{.270} & \textcolor{darkgray}{.264} \\
\textcolor{darkgray}{Ref-based (recall)} & \textcolor{darkgray}{.398} & \textcolor{darkgray}{.380} & \textcolor{darkgray}{.282} & \textcolor{darkgray}{.268} & \textcolor{darkgray}{-.168} & \textcolor{darkgray}{-.164} & \textcolor{darkgray}{.199} & \textcolor{darkgray}{.205} & \textcolor{darkgray}{.248} & \textcolor{darkgray}{.248} \\
Ref-based (F$_1$) & \textcolor{skyblue}{.345} & .304 & .224 & .203 & -.120 & -.135 & .196 & .201 & .259 & .256 \\
\hline
\multicolumn{11}{l}{ Prompt Scorer Gemini-Flash} \\
\hline
Ref-less & .275 & .287 & \textcolor{skyblue}{.256} & \textcolor{orangebrown}{.269} & .027 & .019 & .147 & .146 & .287 & .333 \\
\textcolor{darkgray}{Ref-based (prec)} & \textcolor{darkgray}{.086} & \textcolor{darkgray}{.232} & \textcolor{darkgray}{.105} & \textcolor{darkgray}{-.004} & \textcolor{darkgray}{-.481} & \textcolor{darkgray}{ -.450} & \textcolor{darkgray}{.337} & \textcolor{darkgray}{.181} & \textcolor{darkgray}{.490} & \textcolor{darkgray}{.368} \\
\textcolor{darkgray}{Ref-based (recall)} & \textcolor{darkgray}{.132} & \textcolor{darkgray}{.270} & \textcolor{darkgray}{.103} & \textcolor{darkgray}{.106} & \textcolor{darkgray}{-.577} & \textcolor{darkgray}{-.619} & \textcolor{darkgray}{.330} & \textcolor{darkgray}{.291} & \textcolor{darkgray}{.527} & \textcolor{darkgray}{.531} \\
Ref-based (F$_1$) & .104 & .249 & .104 & -.008 & \textcolor{skyblue}{-.524} & \textcolor{skyblue}{-.521} & \textcolor{skyblue}{.333} & .223 & \textcolor{skyblue}{.508} & \textcolor{skyblue}{.435} \\
\hline
\multicolumn{11}{l}{ Prompt Scorer Llama 3.1} \\
\hline
Ref-less & .297 & .286 & .222 & .210 & -.011 & -.008 & .134 & .116 & .227 & .250 \\
\textcolor{darkgray}{Ref-based (prec)} & \textcolor{darkgray}{.341} & \textcolor{darkgray}{.336} & \textcolor{darkgray}{.178} & \textcolor{darkgray}{.181} & \textcolor{darkgray}{-.120} & \textcolor{darkgray}{-.122} & \textcolor{darkgray}{.080} & \textcolor{darkgray}{.111} & \textcolor{darkgray}{.131} & \textcolor{darkgray}{.172} \\
\textcolor{darkgray}{Ref-based (recall)} & \textcolor{darkgray}{.299} & \textcolor{darkgray}{.313} & \textcolor{darkgray}{.174} & \textcolor{darkgray}{.179} & \textcolor{darkgray}{-.060} & \textcolor{darkgray}{-.048} & \textcolor{darkgray}{.148} & \textcolor{darkgray}{.143} & \textcolor{darkgray}{.175} & \textcolor{darkgray}{.187} \\
Ref-based (F$_1$) & .318 & .324 & .176 & .180 & -.080 & -.069 & .104 & .125 & .150 & .179 \\
\hline
\multicolumn{11}{l}{ Trained Scorer} \\
\hline
Ref-based & .116 & .033 & .091 & .063 & -.133 & .024 & .057 & .059 & .145 & .110 \\
Proxy-ref & .338 & \textcolor{skyblue}{.348} & \textcolor{orangebrown}{.230} & \textcolor{skyblue}{.286} & -.011 & -.057 & .293 & \textcolor{skyblue}{.329} & .298 & .374 \\
\hline
\multicolumn{11}{l}{ Baselines} \\
\hline
RougeL & .150 & .169 & .180 & .190 & .057 & .040 & .124 & .131 & .086 & .099 \\
BLEU & .236 & .184 & .180 & .166 & -.144 & -.039 & .040 & .038 & .107 & .079 \\
Meteor & .229 & .240 & .192 & .191 & -.150 & -.132 & .061 & .064 & .062 & .076 \\
H-METEOR & .005 & -.024 & .076 & .057 & .117 & .025 & .039 & .024 & .008 & .003 \\
\hline
\hline
\multicolumn{11}{l}{\textbf{Weighted Scorer (\(\alpha = 0.5\))}} \\[0.2em]
\hline
GPT-4o / Proxy-ref   & .321 & .323 & .211 & .240 & -.060 & -.077 & .247 & .271 & .297 & .332 \\
Gemini-Pro / Proxy-ref & \textcolor{orangebrown}{.341} & .326 & .227 & .244 & -.066 & -.096 & .245 & .265 & .278 & .315 \\
Gemini-Flash / Proxy-ref & .221 & .299 & .167 & .147 & \textcolor{orangebrown}{-.267} & \textcolor{orangebrown}{-.289} & \textcolor{orangebrown}{.313} & \textcolor{orangebrown}{.276} & \textcolor{orangebrown}{.403} & \textcolor{orangebrown}{.404} \\
Llama 3.1 / Proxy-ref  & .328 & \textcolor{orangebrown}{.336} & .203 & .233 & -.046 & -.063 & .199 & .227 & .224 & .277 \\
\hline
\hline
\end{tabular}}
\caption{\label{table:results_averitec_shared_task_trimmed_full_dimensions} Correlation between human-rated \textbf{AVeriTeC shared task submissions} and proposed scorers, across five quality dimensions. \textcolor{skyblue}{Highest scores per column} are colored blue and \textcolor{orangebrown}{second-highest values} highlighted with brown color. \textcolor{darkgray}{Reference-based precision and recall} scores which are calculated to compute the final reference-based $F_1$ score, are colored in gray.}
\vspace{-1em}
\end{table*}

\section{Evaluation examples}

\begin{table*}[ht]
    \centering
    \scalebox{0.78}{
    \begin{tabular}{p{0.02\linewidth} p{0.13\linewidth} p{0.4\linewidth} p{0.45\linewidth}} \toprule
        Id & Claim & Reference Evidence & Predicted evidence \\\midrule
        602 & Nigeria is the fifth largest receiver of diaspora remittances in the world.
        & \textit{Which countries are the top 5 largest receivers of diaspora remittances in the world?\newline In 2020, the top 5 countries were India, China, Mexico, the Philippines and Egypt. \newline\newline
        Where does Nigeria rank in the list of countries that are the largest receivers of diaspora remittances?\newline Nigeria remains the seventh-largest recipient.\newline\newline} (verdict: refuted)
        & \textit{What is the definition of diaspora remittances?\newline Diaspora remittances are money transfers made by citizens living outside their home country. \newline\newline
        What is the ranking of Nigeria in terms of diaspora remittances?\newline Nigeria is ranked as the sixth largest receiver of remittances in the world.\newline\newline} (verdict: refuted)
    \\\bottomrule
    \end{tabular}}
    \caption{Example from a system submitted to AVeriTeC shared task. Both reference and predicted evidence contradict the claim but represent alternative evidence paths.}
    \label{table:ref_based_evidence_eval_examples}
\end{table*}

\begin{table*}[ht]
    \centering
    \scalebox{0.75}{
    \begin{tabular}{p{0.02\linewidth} p{0.17\linewidth} p{0.38\linewidth} p{0.69\linewidth}} \toprule
        Id & Claim & Reference evidence & Predicted evidence \\\midrule
        602 & Global warming is occurring because of solar winds and changes in Earth's magnetic field.
        & \textit{Does solar wind affect global warming?\newline Most of the particles from solar wind  is reflected back into space by the Earth’s magnetic field. Because the energy does not reach our planet’s surface, it has no measurable influence on surface temperature.\newline \newline
        Do changes in earth's magnetic field impact global warming?\newline There’s no known physical mechanism capable of connecting weather conditions at Earth’s surface with electromagnetic currents in space.\newline\newline
        What is causing global warming?\newline Scientists attribute the global warming trend observed since the mid-20th century to the human expansion of the ""greenhouse effect"" — warming that results when the atmosphere traps heat radiating from Earth toward space.}
        & \textit{What causes global warming?\newline Global warming is primarily caused by human activity, specifically the emission of carbon dioxide (CO2) and other greenhouse gases. These gases trap excess heat on Earth, leading to an increase in the planet's average temperature. \newline\newline
        What is the role of solar winds in global warming?\newline Solar winds, consisting of charged particles emitted by the sun, do not contribute to global warming. Earth's magnetic field prevents most of these particles from interacting with Earth. Changes in the sun's behavior cannot explain modern global warming. The warming trend is caused by human-emitted greenhouse gases trapping excess heat on the Earth.\newline\newline
        Can changes in Earth's magnetic field cause global warming?\newline No, changes in Earth's magnetic field do not have an impact on global temperatures.\newline\newline
        What is the scientific consensus on the cause of global warming?\newline
        The scientific consensus on the cause of global warming is that it is primarily caused by human activities, particularly the burning of fossil fuels which leads to the emission of greenhouse gases into the Earth's atmosphere.
        % \newline\newline
        % What is the primary cause of global warming?\newline The primary cause of global warming is human activity, particularly the burning of fossil fuels which releases CO2 into the atmosphere. This acts like a blanket, trapping heat that would otherwise escape into space and causing global warming. The energy sector is responsible for almost three-quarters of the emissions that have caused global temperatures to warm by 1.1°C since pre-industrial times.
        [...]}
    \\\bottomrule
    \end{tabular}}
    \caption{Example for predicted evidence with repetitions.}
    \label{table:repetition_evidence_eval_examples}
\end{table*}

\begin{table*}[ht]
    \centering
    \scalebox{0.77}{
    \begin{tabular}{p{0.02\linewidth} p{0.17\linewidth} p{0.42\linewidth} p{0.6\linewidth}} \toprule
        Id & Claim & Reference evidence & Predicted evidence \\\midrule
        56 & Scotland is spending more on health per head than the rest of the UK. 
        & \textit{How much does the UK spend on healthcare?\newline The UK as a whole spends the equivalent of one quarter of total public expenditure and £2,972 per capita in 2017.\newline\newline
        How much does Scotland spend on healthcare?\newline Scotland spends 21 percent or £8bn on healthcare, per person £2,507.\newline\newline
        How much does Northern Ireland spend on healthcare?\newline 22 percent or £5bn and £2,616 per person.\newline\newline
        How much does Wales spend on healthcare?\newline Wales spends 23 percent or £13.7bn and £2,546 per person.\newline\newline
        How much does England spend on healthcare?\newline 25 percent or £136.6bn and £2,427 per person.}
        & \textit{What is the health spending per head in Scotland?\newline In 2013/14, Scotland spent £2,187 per head in real terms.\newline\newline
        What is the health spending per head in England?\newline In 2013/14, England spent £2,028 per head in real terms.\newline\newline
        How does the health spending per head in Scotland compare to the rest of the UK?\newline Scotland continues to spend the most per head on health compared to the rest of the UK.\newline\newline
        What is the trend in health spending per head in the UK since 2003/04?\newline All four countries of the UK spend more per head on health now than they did in 2003/04, but the trend has flattened since 2009/10.\newline\newline
        What is the health spending per head in Northern Ireland?\newline In 2013/14, Northern Ireland spent £2,158 per head in real terms.\newline\newline
        What is the health spending per head in Wales?\newline In 2013/14, Wales spent £2,026 per head in real terms. [...]
        % \newline\newline
        % What is the average annual growth rate of health spending per head in England?\newline The average annual growth rate of health spending per head in England is 2.5\%.\newline\newline
        % What is the average annual growth rate of health spending per head in Wales?\newline The average annual growth rate of health spending per head in Wales is 1.3\%. \newline\newline
        % What is the health spending per capita in Scotland according to ExpatAssure?\newline The total expenditure on health per capita and per year in Scotland is £2,160.[...]
        }
    \\\bottomrule
    \end{tabular}}
    \caption{Example from systems submitted to AVeriTeC shared task~\citep{schlichtkrull2024automatedverificationtextualclaims} extracted from manual evaluation of proxy-reference evaluation.}
    \label{table:proxy_reference_evidence_eval_examples}
\end{table*}

\begin{table*}[ht]
    \centering
    \scalebox{0.9}{
    \begin{tabular}{p{0.02\linewidth} p{0.18\linewidth} p{0.31\linewidth} p{0.48\linewidth}} \toprule
        Id & Claim & Facts & Predicted evidence \\\midrule
        20 & France24 encouraging Niger Delta militants to fight for their right to equity in resource sharing. & \textit{1. France24 is encouraging Niger Delta militants.\newline\newline 2. The encouragement is to fight for their right to equity in resource sharing.} 
        & \textit{Since groups in the Niger Delta could not be mobilising simply for the sake of doing so, the insight that these analysts attempt to proffer is that the Niger Delta mobilisation must be for the maximisation of sectional interests, with the non-producing ethnic groups a target of their grievance.} \\ \hline
        593 & Antifa vandalized the home of Missouri senator Josh Hawley in 4th of January 2021 
        & \textit{1. Antifa vandalized the home of Missouri senator Josh Hawley.\newline\newline 2. The vandalism occurred on January 4, 2021.}
        % “We walked very nicely down the sidewalk to his house. We had one megaphone, that’s it,” Nadine Bock, a member of ShutDownDC who took part in the vigil, told AFP by phone. “We chanted slogans around respecting the election, respecting democracy, following the constitution. We did not pound on his door,” she added.Hawley’s wife briefly came outside holding her baby, exchanged a few words with the protesters and went back inside. Bock said that protesters did not utter any threats, while Hawley said that they “screamed threats.” Is this backed up by evidence? Yes The entire event was streamed live on YouTube, and the footage corroborates Bock and Vazquez’s versions.
        & \textit{Was there any evidence linking Antifa to the vandalism of Missouri senator Josh Hawley's home on January 4, 2021?\newline In Newark, New Jersey, police and FBI investigators had identified “a probable ‘Antifa’ related individual,” who was arrested for possessing a knife, a hatchet, and a jar of gasoline. \newline\newline
        Where was Senator Josh Hawley residing at the time of the alleged Antifa vandalism on January 4, 2021?\newline who was living in Omaha. Antifa Nebraska printed thousands. [...]}
        \\ \hline
        648 & Biden won fewer counties than Trump and Obama but more votes, suggesting election fraud in 2020.
        & \textit{1. Biden won fewer counties than Trump.\newline\newline 2. Biden won fewer counties than Obama.\newline\newline 3. Biden won more votes than Trump.\newline\newline 4. Biden won more votes than Obama.\newline\newline 5. This suggests election fraud in 2020.}
        % How many votes did Obama get in his Presidential election? 69,498,516. Is there a correlation between the amount of counties and the amount of votes? 
        % Answer: According to a 2017 U.S. Census Bureau report, more than half of the country's population lives in less than 5\% percent of counties — just 143 heavily populated counties. Less than half the population is spread out across the other 2,999 counties
        
        % So the total vote count can vary significantly depending on the population of the counties won.
        & \textit{Did the increase in votes for Biden suggest any irregularities in the 2020 election? The claim that Biden won fewer counties than Trump and Obama but more votes doesn't necessarily suggest election fraud.It is possible for a candidate to win more votes by appealing to a broader base of voters in specific key areas, regardless of the number of counties won. Such an analysis requires a comprehensive understanding of the electoral process, including voter turnout, population distribution, and voting patterns. Without specific evidence of fraud, the claim lacks substantial support.}
    \\\bottomrule
    \end{tabular}}
    \caption{Examples from systems submitted to AVeriTeC shared task. The second column depicts \textit{Facts} that are have been extracted from the respective claim for reference-less evidence evaluation.}
    \label{table:ref_less_evidence_eval_examples}
\end{table*}

\section{Prompt templates}

\begin{figure*}[h]
    \centering
    \resizebox{0.88\textwidth}{!}{%
    \includegraphics{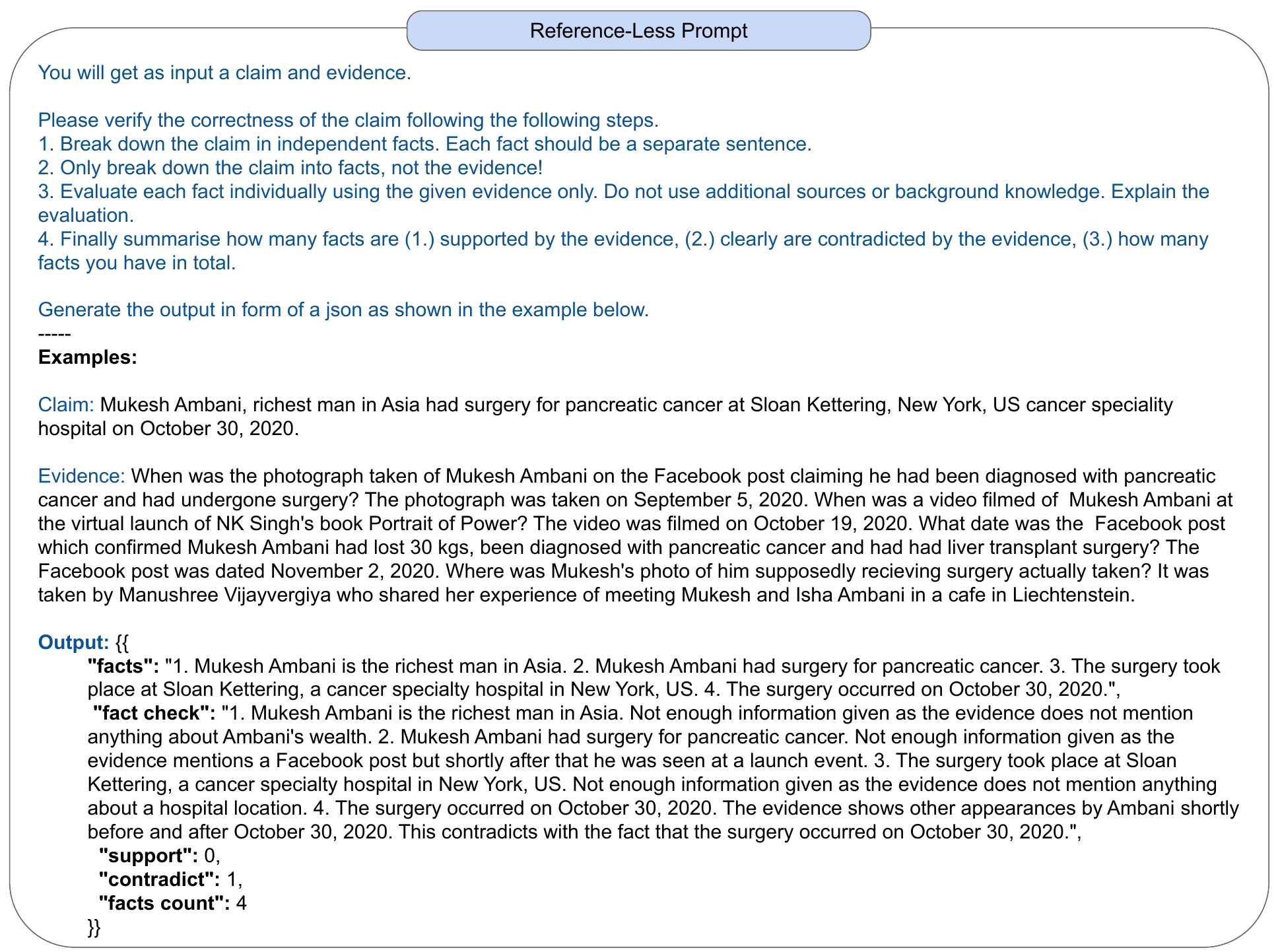}
    }
    \caption{Prompt we used for the Ev\textsuperscript{2}R reference-less prompt scorer. Given as input the claim and retrieved evidence, the scorer decomposes the claim and evaluates the resulting facts against the evidence data.}
    \label{fig:prompt_reference_less_component}
    % \vspace{-1em}
\end{figure*}

\begin{figure*}[h]
    \centering
    \resizebox{0.8\textwidth}{!}{%
    \includegraphics{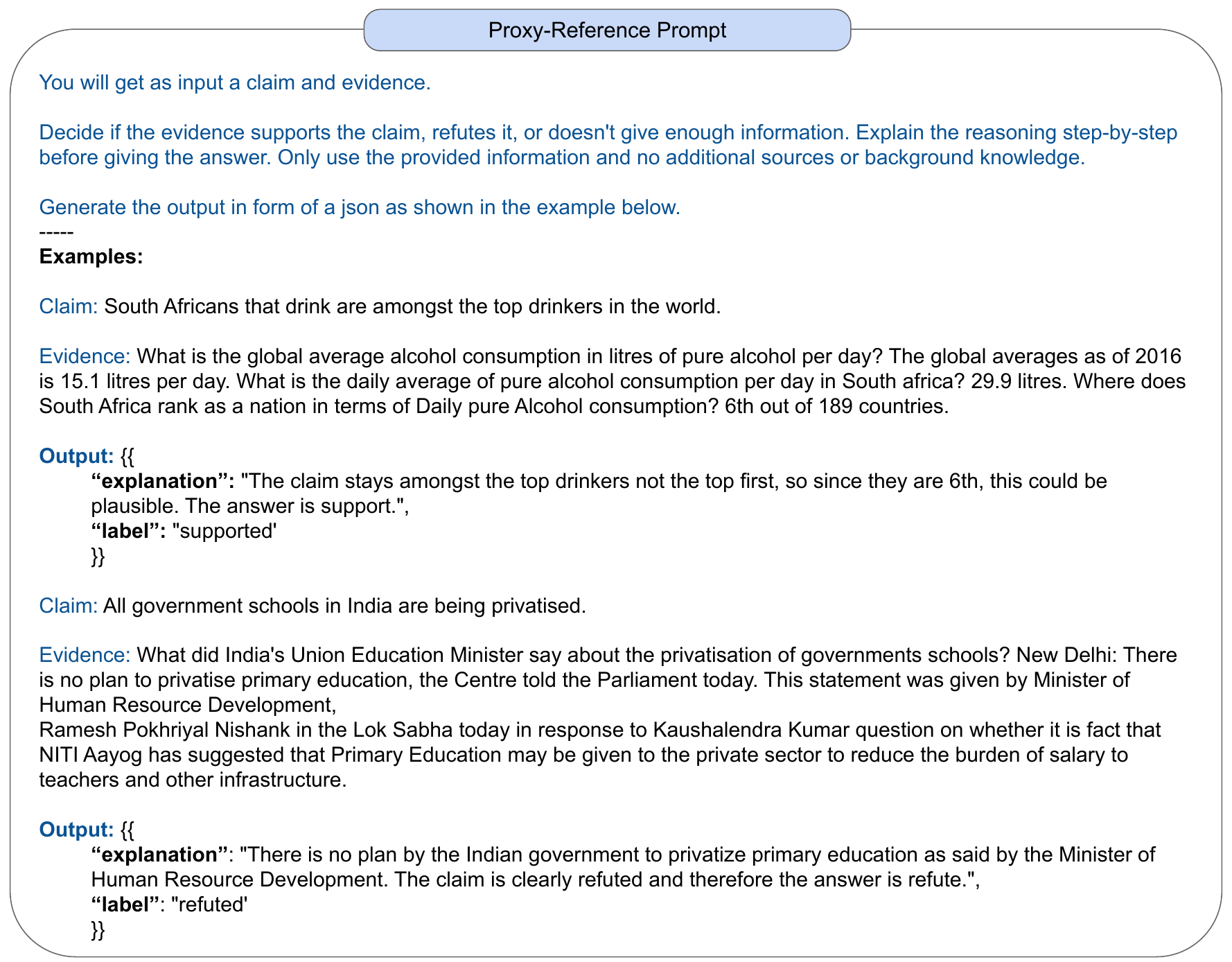}
    }
    \caption{Prompt we used for the Ev\textsuperscript{2}R proxy-reference prompt. We instruct the backbone LLM to reason over the retrieved evidence to predict a verdict label is assessed against the gold verdict label we use as proxy reference.}
    \label{fig:prompt_proxy_reference_component}
    % \vspace{-1em}
\end{figure*}

\begin{figure*}[h]
    \centering
    \resizebox{1.0\textwidth}{!}{%
    \includegraphics{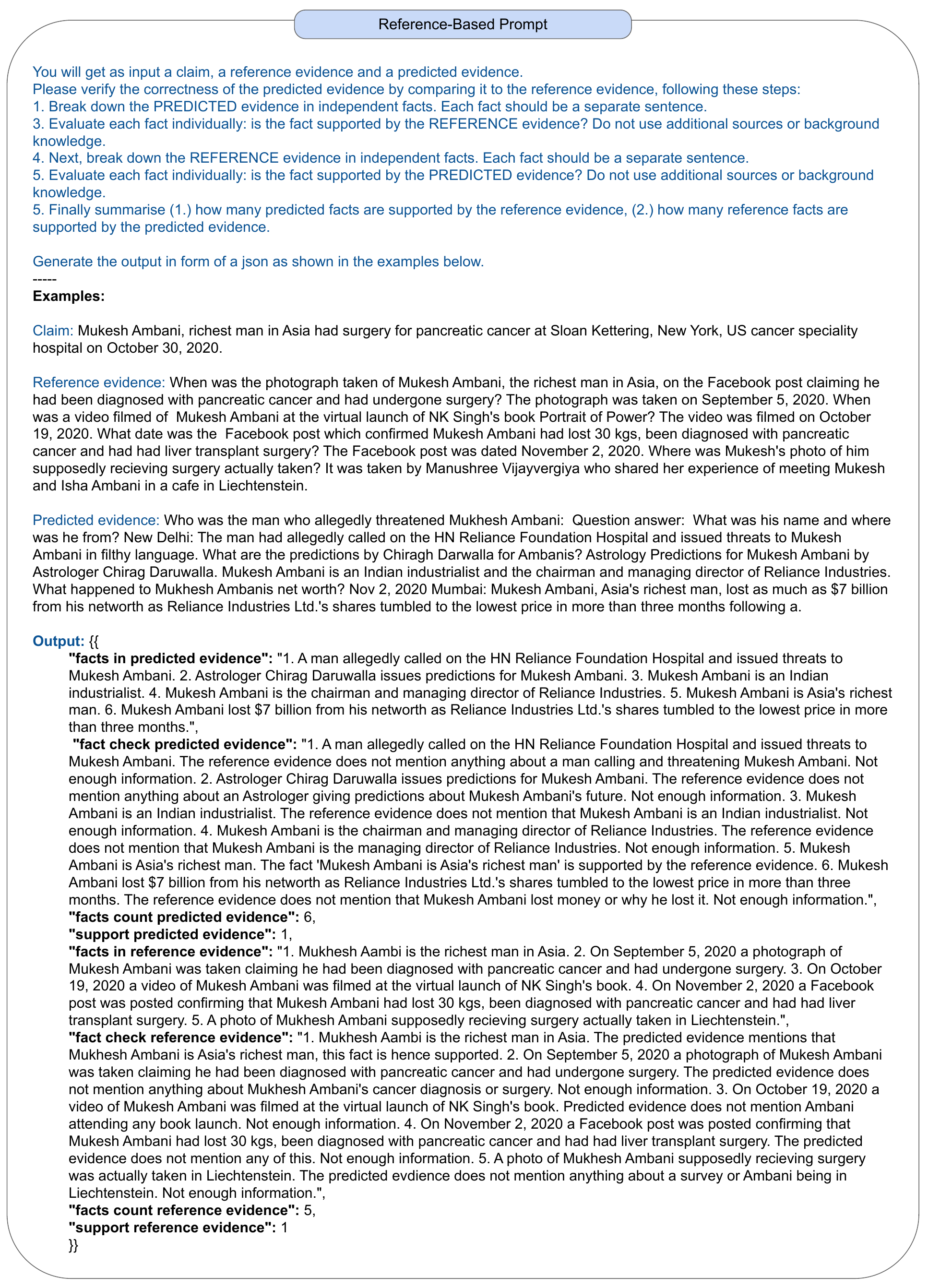}
    }
    \caption{Prompt we used for the Ev\textsuperscript{2}R reference-based prompt scorer. Both the reference evidence and the retrieved evidence are decomposed into atomic facts before assessing them against each other.}
    \label{fig:prompt_reference_based_component}
    % \vspace{-1em}
\end{figure*}

\section{Human evaluation}

\begin{figure*}[ht]
    \centering
    \includegraphics[scale=0.6]{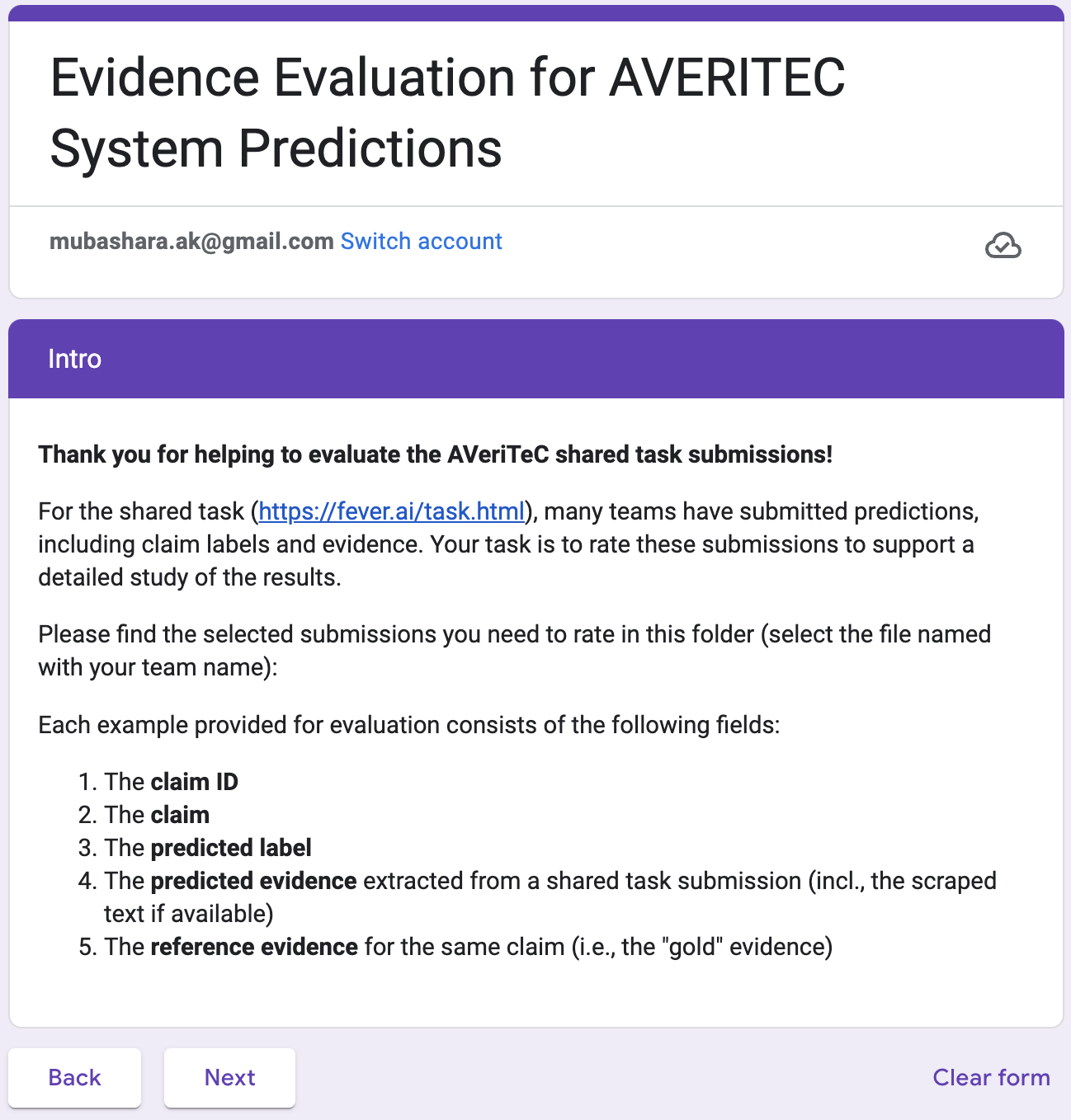}
    \caption{Platform for human evaluation of retrieved evidence from systems submitted to the AveriTeC shared task.}
    \label{fig:test1}
\end{figure*}

\begin{figure*}[ht]
    \centering
    \includegraphics[scale=0.75]{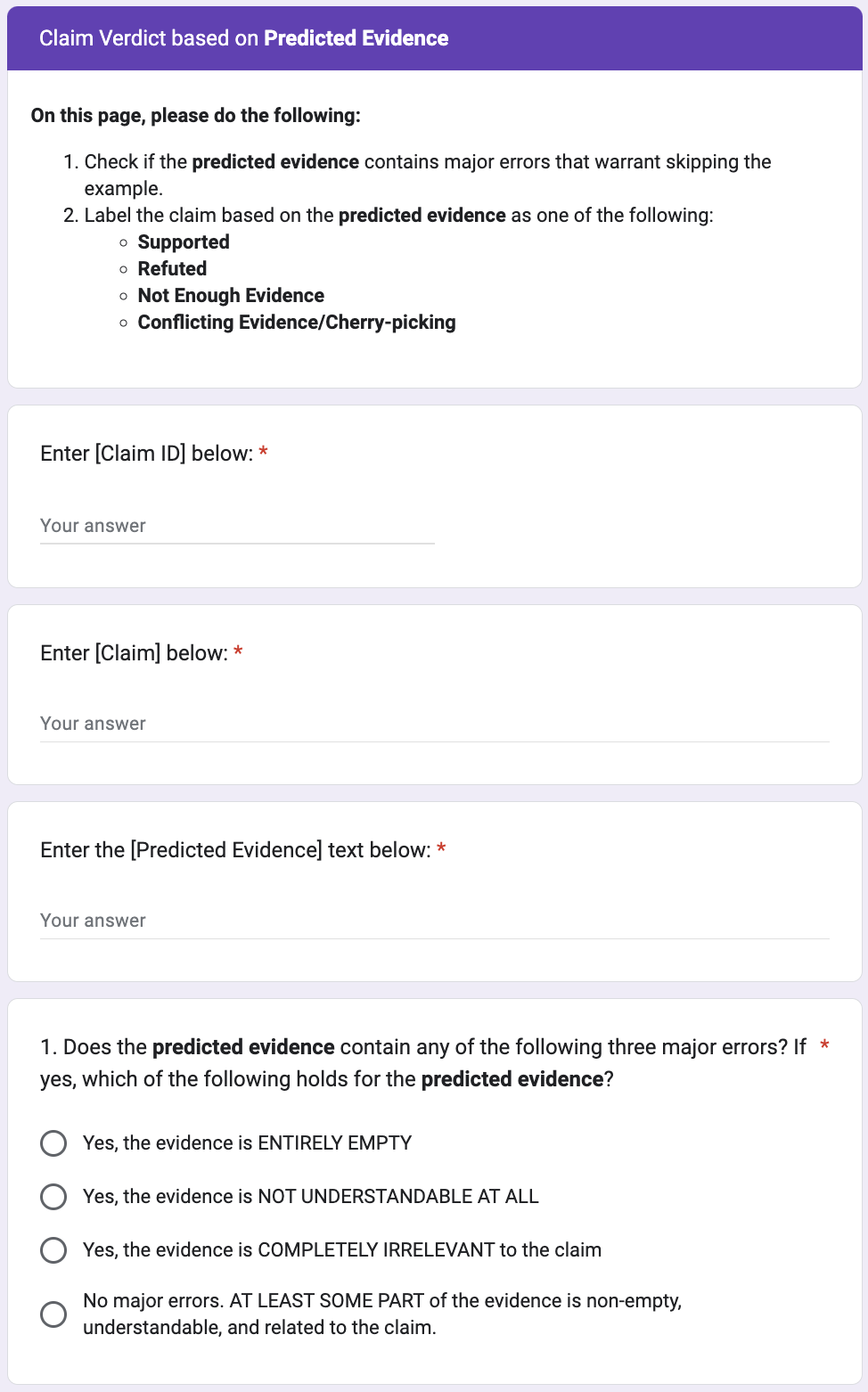}
    \caption{Platform for human evaluation of retrieved evidence from systems submitted to the AveriTeC shared task.}
    \label{fig:test2}
\end{figure*}

\begin{figure*}[ht]
    \centering
    \includegraphics[scale=0.75]{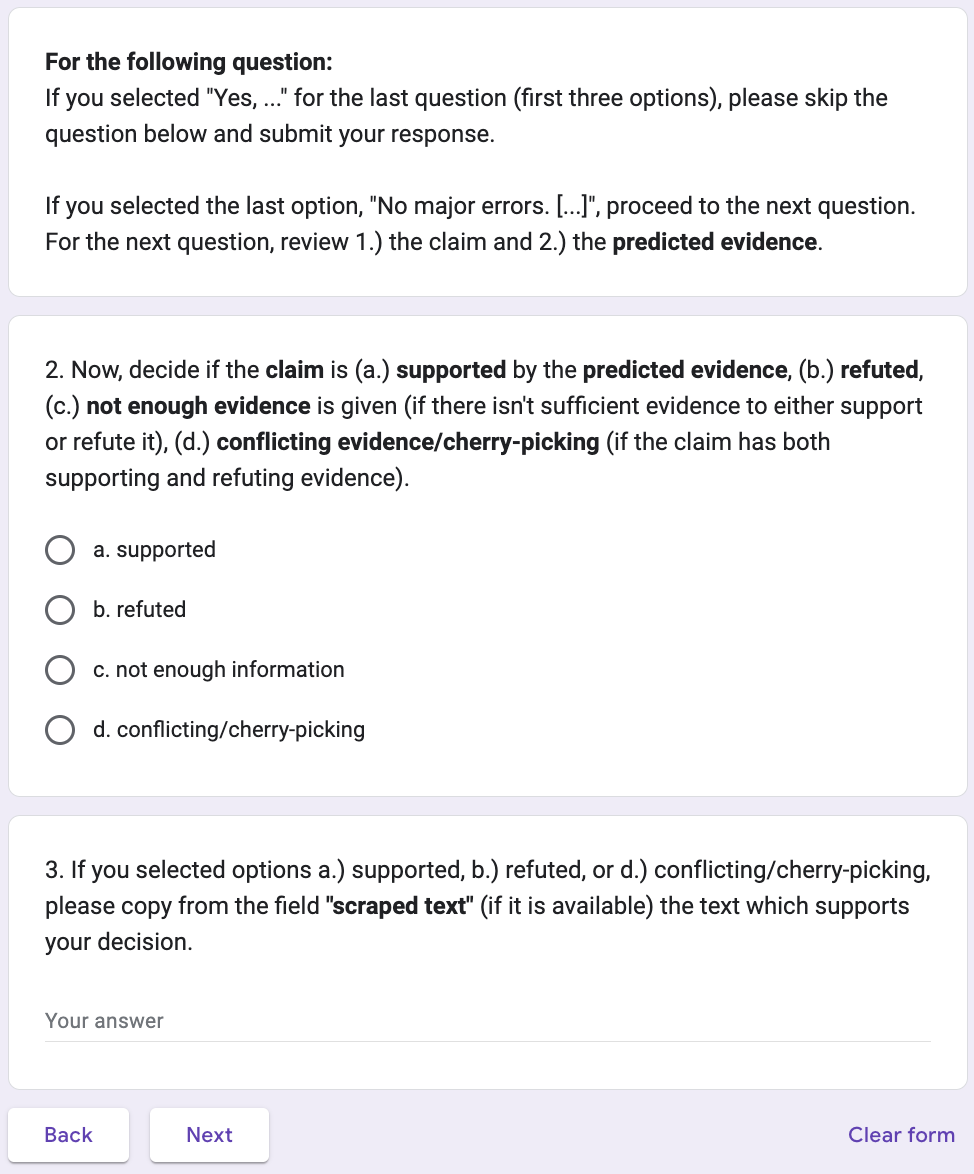}
    \caption{Platform for human evaluation of retrieved evidence from systems submitted to the AveriTeC shared task.}
    \label{fig:test3}
\end{figure*}

\begin{figure*}[ht]
    \centering
    \includegraphics[scale=0.6]{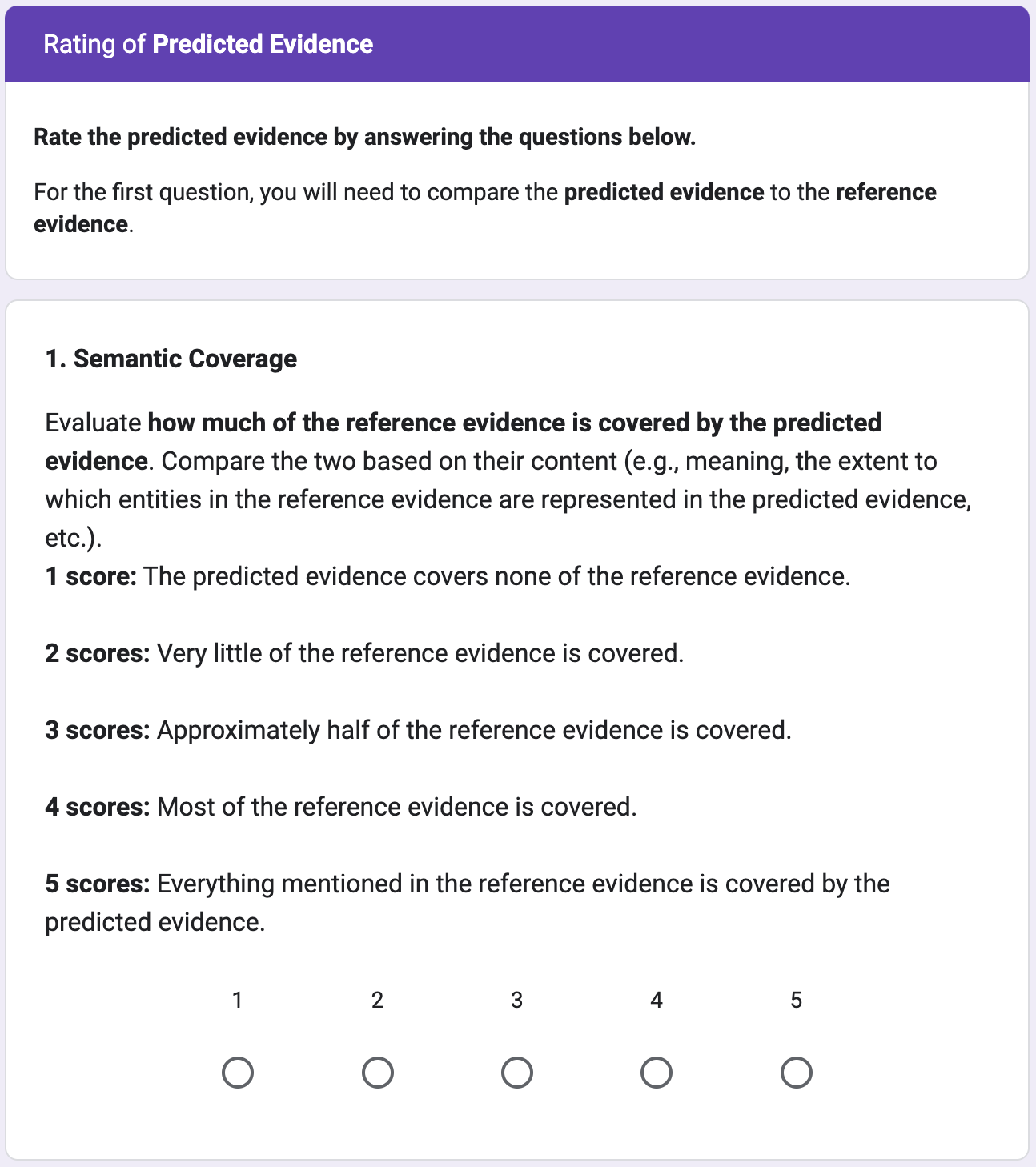}
    \caption{Platform for human evaluation of retrieved evidence from systems submitted to the AveriTeC shared task.}
    \label{fig:test4}
\end{figure*}

\begin{figure*}[ht]
    \centering
    \includegraphics[scale=0.6]{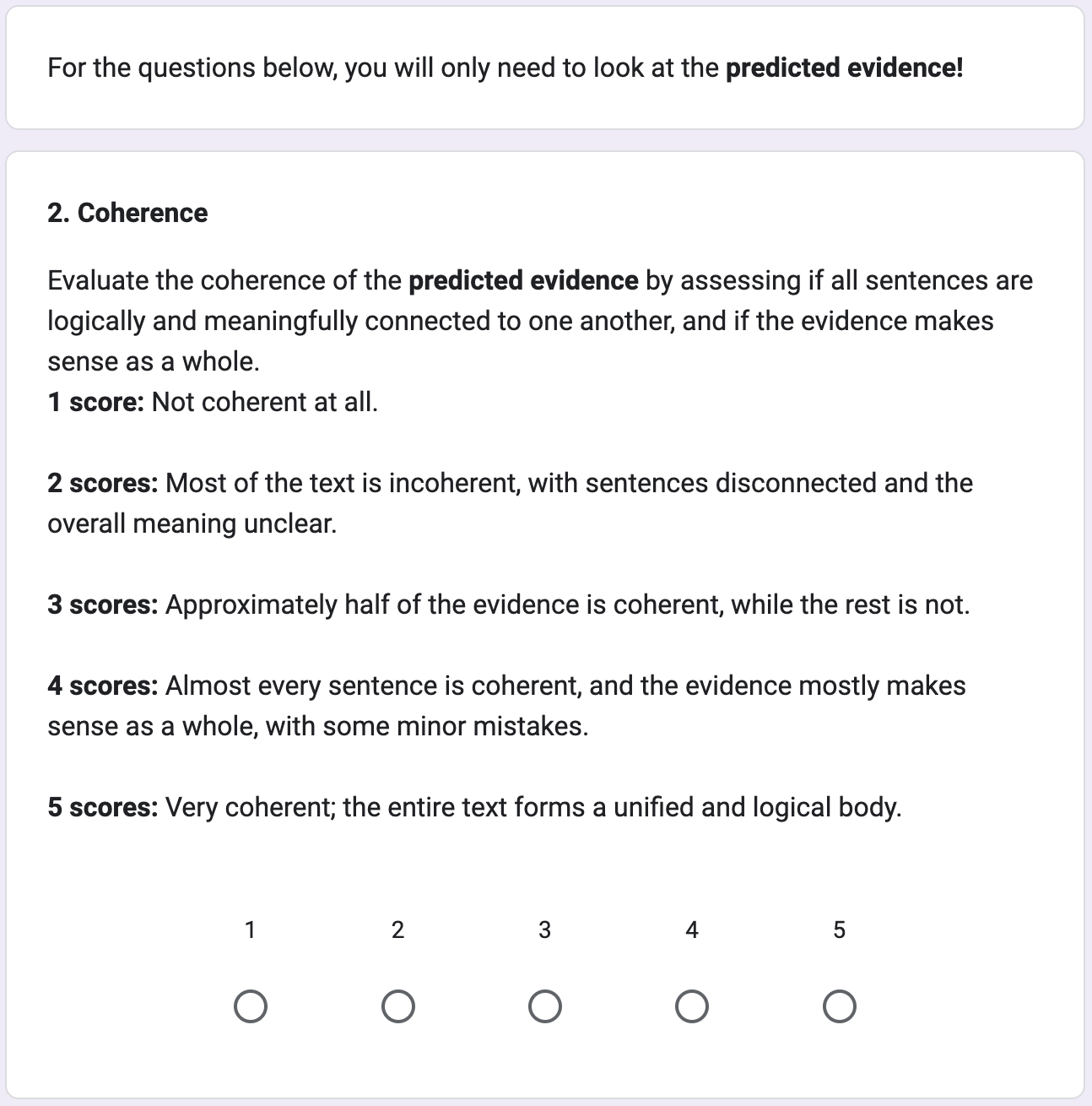}
    \caption{Platform for human evaluation of retrieved evidence from systems submitted to the AveriTeC shared task.}
    \label{fig:test5}
\end{figure*}

\begin{figure*}[ht]
    \centering
    \includegraphics[scale=0.6]{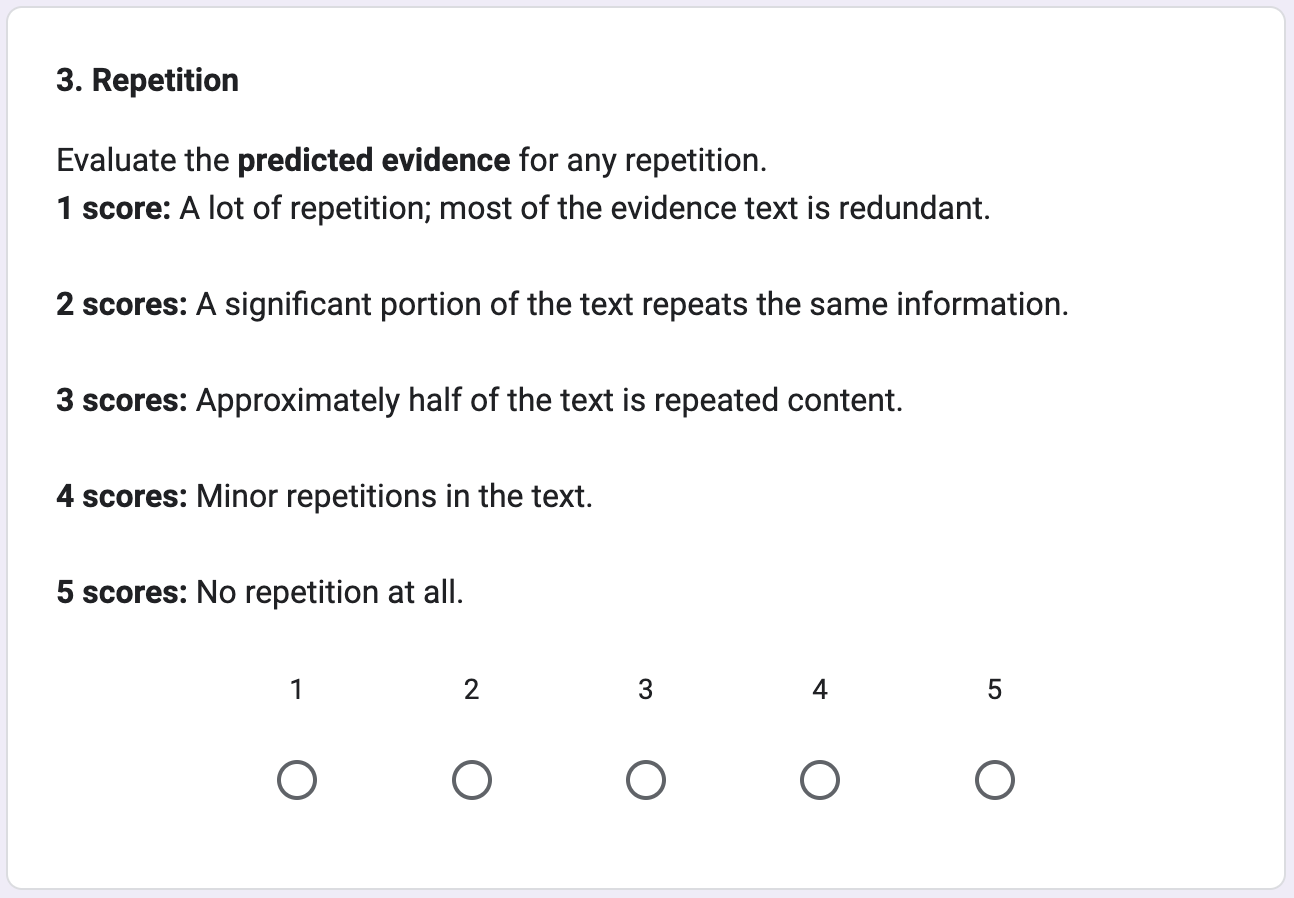}
    \caption{Platform for human evaluation of retrieved evidence from systems submitted to the AveriTeC shared task.}
    \label{fig:test6}
\end{figure*}

\begin{figure*}[ht]
    \centering
    \includegraphics[scale=0.6]{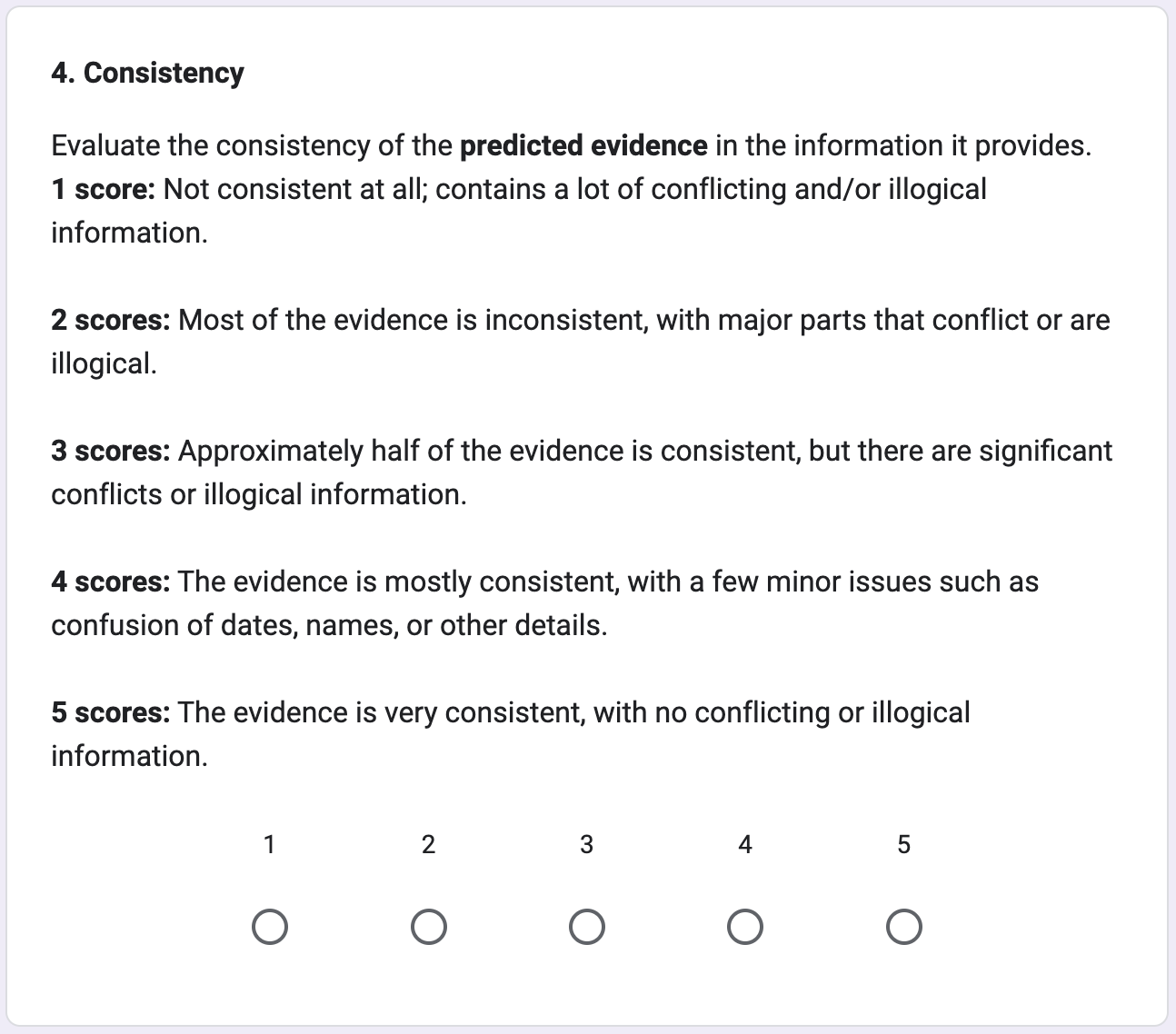}
    \caption{Platform for human evaluation of retrieved evidence from systems submitted to the AveriTeC shared task.}
    \label{fig:test7}
\end{figure*}

\begin{figure*}[ht]
    \centering
    \includegraphics[scale=0.6]{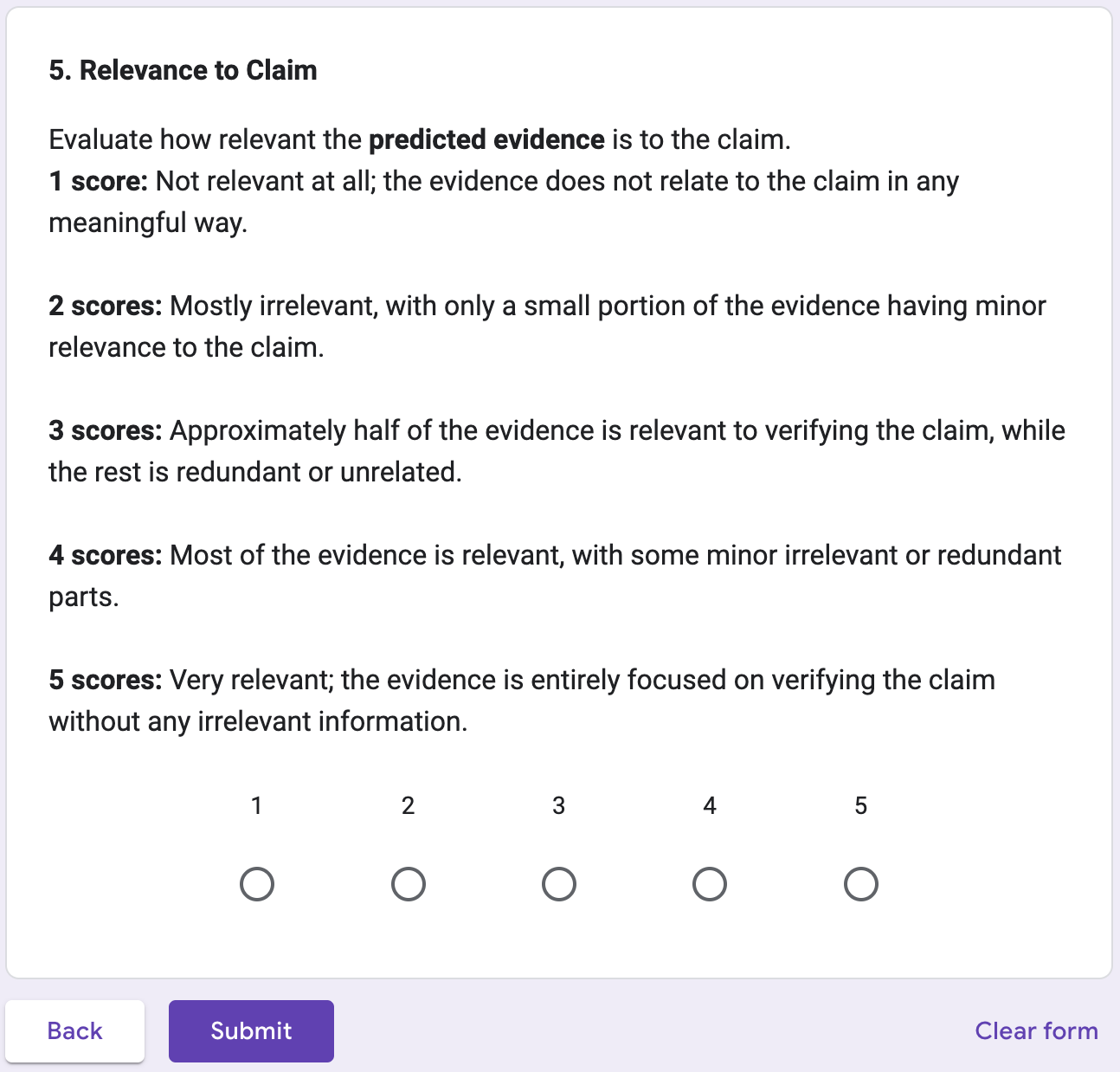}
    \caption{Platform for human evaluation of retrieved evidence from systems submitted to the AveriTeC shared task.}
    \label{fig:test8}
\end{figure*}

% \fi
\begin{table*}[!htbp]
\centering
\scalebox{0.81}{
\begin{tabular}{l | c c | c c }
\hline
\textbf{Scorer Component} & \multicolumn{2}{c|}{\textbf{Spearman $\rho$}} & \multicolumn{2}{c}{\textbf{Pearson $r$}} \\
 & Correlation & p-value & Correlation & p-value \\
\hline
\multicolumn{5}{l}{\bf FEVER} \\
\hline
Gemini Flash Precision & 0.139 & 0.0018 & 0.119 & 0.0080 \\
Gemini Flash Recall & 0.139 & 0.0019 & 0.105 & 0.0193 \\
Gemini Pro Precision & 0.139 & 0.0019 & 0.118 & 0.0082 \\
Gemini Pro Recall & 0.176 & 7.86e-05 & 0.143 & 0.0013 \\
GPT-4o Precision & 0.140 & 0.0016 & 0.129 & 0.0038 \\
GPT-4o Recall & 0.146 & 0.0011 & 0.114 & 0.0107 \\
Proxy Score & 0.375 & 3.48e-18 & 0.393 & 6.29e-20 \\
\hline
\multicolumn{5}{l}{\bf VitaminC} \\
\hline
Gemini Flash Precision & 0.202 & 5.10e-06 & 0.198 & 8.49e-06 \\
Gemini Flash Recall & 0.232 & 1.48e-07 & 0.227 & 2.87e-07 \\
Gemini Pro Precision & 0.222 & 5.27e-07 & 0.216 & 1.14e-06 \\
Gemini Pro Recall & 0.253 & 9.65e-09 & 0.246 & 2.66e-08 \\
GPT-4o Precision & 0.203 & 4.53e-06 & 0.200 & 6.67e-06 \\
GPT-4o Recall & 0.229 & 2.37e-07 & 0.228 & 2.70e-07 \\
Proxy Score & 0.250 & 1.53e-08 & 0.273 & 5.63e-10 \\
\hline
\multicolumn{5}{l}{\bf AVeriTeC} \\
\hline
Gemini Flash Precision & -0.081 & 0.803 & -0.034 & 0.00916 \\
Gemini Flash Recall & -0.051 & 0.875 & 0.030 & 0.00925 \\
Gemini Pro Precision & 0.272 & 1.42e-05 & 0.226 & 0.00034 \\
Gemini Pro Recall & 0.282 & 6.97e-06 & 0.269 & 1.86e-05 \\
GPT-4o Precision & 0.283 & 1.68e-06 & 0.260 & 1.17e-05 \\
GPT-4o Recall & 0.316 & 7.31e-08 & 0.284 & 1.47e-06 \\
Proxy Score & 0.475 & 1.29e-15 & 0.450 & 5.99e-14 \\
\hline
\end{tabular}}
\caption{\label{table:results_significance} Correlation between reference and predicted scores for the evaluation dimension `verdict agreement' on FEVER, VitaminC, and AVeriTeC.}
\vspace{-1em}
\end{table*}

\end{document}